\documentclass[10pt,twocolumn,letterpaper]{article}
\usepackage[pagenumbers]{cvpr} % To force page numbers, e.g. for an arXiv version
\usepackage{graphicx}
\usepackage{amsmath}
\usepackage{amssymb}
\usepackage{booktabs}
\usepackage{algorithmic}
\usepackage{csquotes}
\usepackage{amsmath}
 \usepackage[ruled]{algorithm2e}               
\usepackage[pagebackref,breaklinks,colorlinks]{hyperref}

\usepackage[capitalize]{cleveref}
\crefname{section}{Sec.}{Secs.}
\Crefname{section}{Section}{Sections}
\Crefname{table}{Table}{Tables}
\crefname{table}{Tab.}{Tabs.}

\def\confName{CVPR} 
\def\confYear{2023}

\begin{document}

\title{Towards Trustable Skin Cancer Diagnosis via Rewriting Model's Decision}

% \author{Siyuan Yan\\
% Faculty of Engineering, Monash University\\
% \and

\author{
Siyuan Yan$^{1,2}$\quad
Zhen Yu$^{1,2}$\quad
Xuelin Zhang$^{1,2}$ \quad
Dwarikanath Mahapatra$^{4}$\quad\\
Shekhar S. Chandra$^{3}$\quad
Monika Janda$^3$\quad
Peter Soyer$^3$\quad
Zongyuan Ge$^{1,2}$\thanks{Corresponding author: Zongyuan Ge \emph{(zongyuan.ge@monash.edu)}}\quad\\
$^1$ Monash University\quad
$^2$ Monash Medical AI Group 
$^3$ The University of Queensland \quad\\
$^4$ Inception Institute of AI, Abu Dhabi, UAE \quad\\
% {\tt \small \href{http://dpfan.net/UCNet/}{http://dpfan.net/UCNet/}}\\
}

\maketitle

%%%%%%%%% ABSTRACT
\begin{abstract}
     Deep neural networks have demonstrated promising performance on image recognition tasks. However, they may heavily rely on confounding factors, using irrelevant artifacts or bias within the dataset as the cue to improve performance. When a model performs decision-making based on these spurious correlations, it can become untrustable and lead to catastrophic outcomes when deployed in the real-world scene. In this paper, we explore and try to solve this problem in the context of skin cancer diagnosis. We introduce a human-in-the-loop framework in the model training process such that users can observe and correct the model's decision logic when confounding behaviors happen. Specifically, our method can automatically discover confounding factors by analyzing the co-occurrence behavior of the samples. It is capable of learning confounding concepts using easily obtained concept exemplars. By mapping the black-box model's feature representation onto an explainable concept space, human users can interpret the concept and intervene via first order-logic instruction. We systematically evaluate our method on our newly crafted, well-controlled skin lesion dataset and 
    several public skin lesion datasets. Experiments show that our method can effectively detect and remove confounding factors from datasets without any prior knowledge about the category distribution and does not require fully annotated concept labels. We also show that our method enables the model to focus on clinical-related concepts, improving the model's performance and trustworthiness during model inference. 

    \end{abstract}

%%%%%%%%% BODY TEXT
\section{Introduction} 
\label{sec:intro}
Deep neural networks have achieved excellent performance on many visual recognition tasks \cite{resnet,deepderm,inception}. Meanwhile, there are growing concerns about the trustworthiness of the model's black-box decision-making process. 
In medical imaging applications, one of the major issues is deep learning models' confounding behaviors using irrelevant artifacts (\ie~rulers, dark corners) or bias (\ie~image backgrounds, skin tone) as the cue to make the final predictions. These spurious correlations in the training distribution can make models fragile when novel testing samples are presented. 
Therefore, transparency of decision-making and human-guided bias correction will significantly increase the reliability and trustworthiness of model deployments in a life-critical application scenario like cancer diagnosis.

\begin{figure}[!t]
  \centering
%   \fbox{\rule{0pt}{2in} \rule{0.9\linewidth}{0pt}}
   \includegraphics[width=0.9\linewidth]{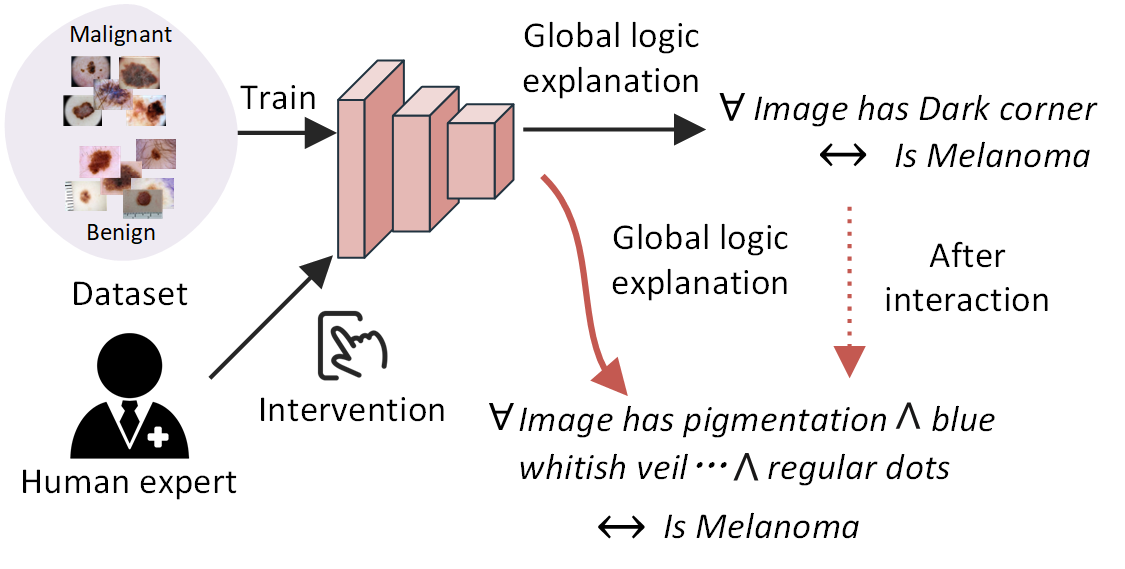}
     \vspace{-2mm}
   \caption{Our method allows people to correct the model's confounding behavior within the skin lesion training set via rewriting the model's logic.}
     \vspace{-4mm}
   \label{fig1}
   \end{figure}
   
For instance, due to the nature of dermatoscopic images, the skin cancer diagnosis often involves confounding factors \cite{bias1,bias2,bias3,bias4}, \ie, dark corners, rulers, and air pockets. Bissoto \etal \cite{bias1} shows that deep learning models trained on common skin lesion datasets can be biased. Surprisingly, they found the model can reach an AUC of 73\%, even though lesions in images are totally occluded. To better understand the issue of confounding behaviors, we illustrate a motivating example. Fig.~\ref{fig1} shows a representative confounding factor ``dark corners" in the \textit{ISIC2019-2020} \cite{isic2019,isic2020} dataset, where the presence of dark corners is much higher in the melanoma images than in benign images. A model trained on this dataset tends to predict a lesion as melanoma when dark corners appear. This model is undoubtedly untrustable and is catastrophic for deployment in real clinical practice. To deal with this nuisance and improve the model's trustworthiness, an ideal method would be: i ) the model itself is explainable so that humans can understand the prediction logic behind it. and ii) the model has an intervening mechanism that allows humans to correct the model once confounding behavior happens. Through this way, the model can avoid dataset biasing issues and rely on expected clinical rules when needed to diagnose. 

% To deal with such nuisance and improve the models' trustworthiness, an ideal method would be for the model to be self-explainable, reveal the prediction logic behind it, and have a mechanism to allow physicians to intervene in the decision-making process and correct the model. In this way, the model can avoid dataset biasing issues and rely on expected clinical rules when needed to diagnose. 

In this work, our overall goal is to improve model transparency as well as develop a mechanism that human-user can intervene the model training process when confounding factors are observed.
%Motivated by these, the objective of this paper is to explain any model's behavior and remove its confounding behavior by human-user, making the model trustable. 
The major difference between our proposed solution and existing works \cite{rrr_nmi,rrr,CBM,Nesy} are: 
(1) For \textbf{model visualization}, we focus on generating human-understandable textual concepts rather than pixel-wise attribution explanations like Class Activation Maps \cite{cam}. 
(2) For \textbf{concept learning}, we do not hold any prior knowledge about the confounding concepts within the dataset. This makes our problem set-up more realistic and practical than the existing concept-based explanation or interaction \cite{CBM,Nesy} framework where fully-supervised concept labels are required. 
%which only works when rich pre-defined concept labels exist. 
(3) For \textbf{method generalization}, our method can be applied on top of any deep models.

Therefore, we propose a method that is capable of learning confounding concepts with easily obtained concept examplars. This is realized via clustering model's co-occurrence behavior based on spectral relevance analysis \cite{spay} and concept exemplars learning based on concept activation vector (CAVs) \cite{cav}. Also, a concept-based space represented by CAVs is more explainable than the feature-based counterparts. 

\begin{figure}[!t]
  \begin{center}
  \begin{tabular}{ c@{ } c@{ } c@{ } c@{ } c@{ }}
  {\includegraphics[width=0.180\linewidth]{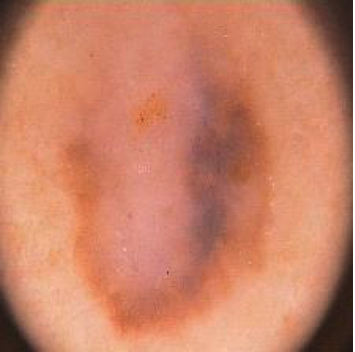}} &
\includegraphics[width=0.180\linewidth]{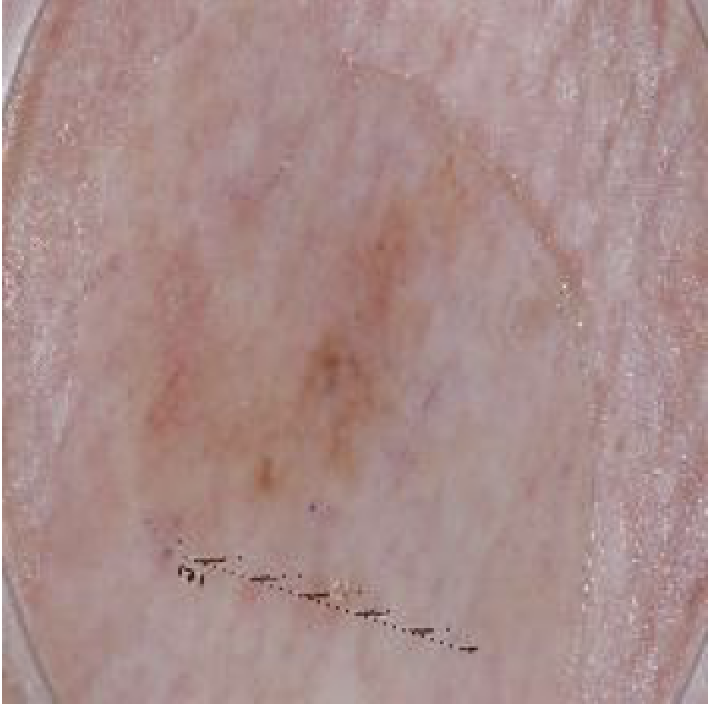}&
  {\includegraphics[width=0.180\linewidth]{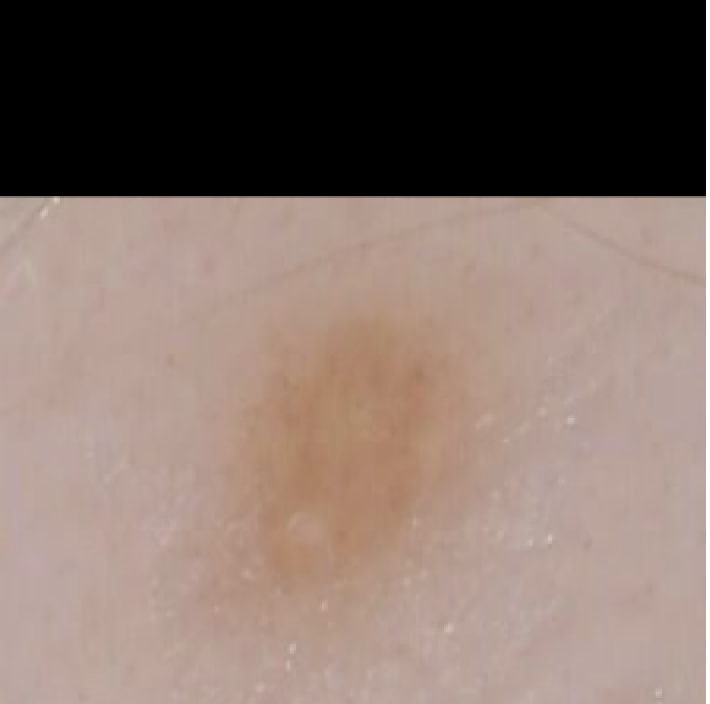}}&
  {\includegraphics[width=0.180\linewidth]{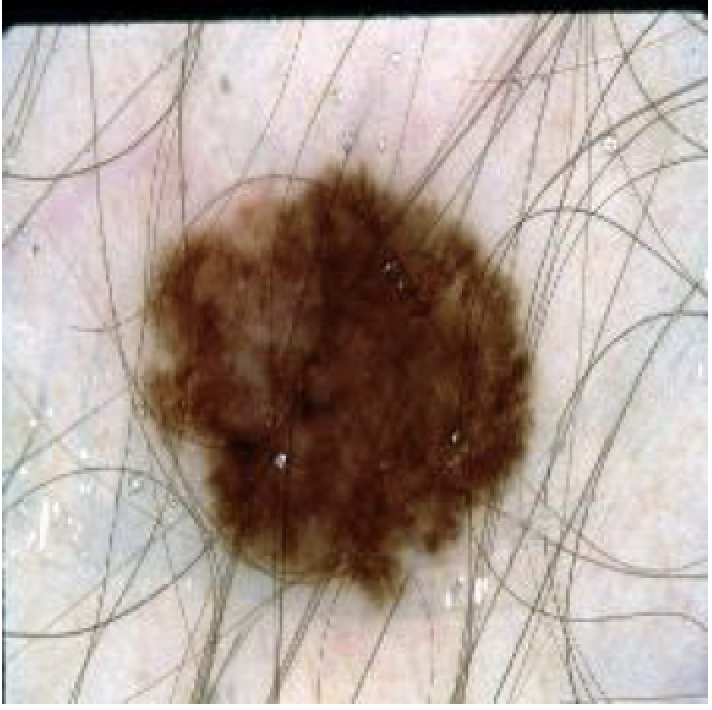}} &
  {\includegraphics[width=0.180\linewidth]{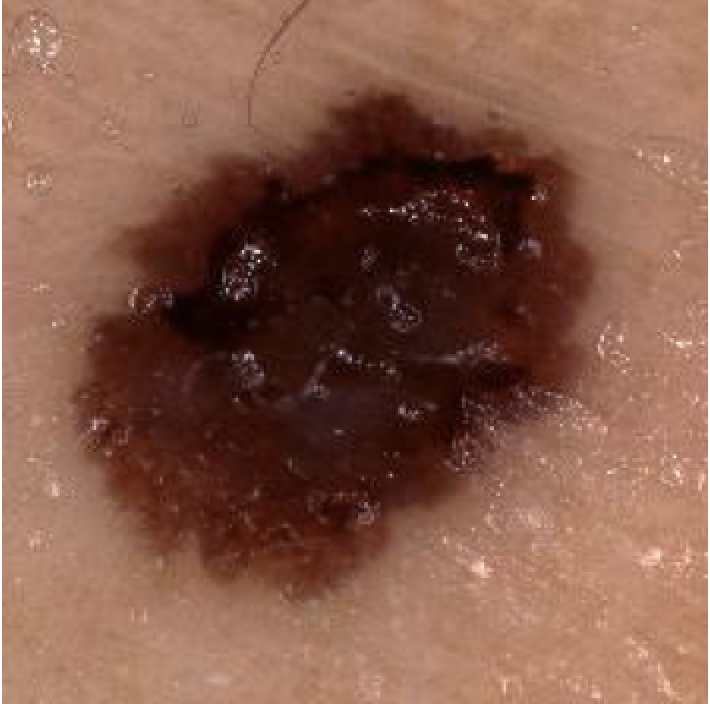}} 
  \\
  {\includegraphics[width=0.180\linewidth]{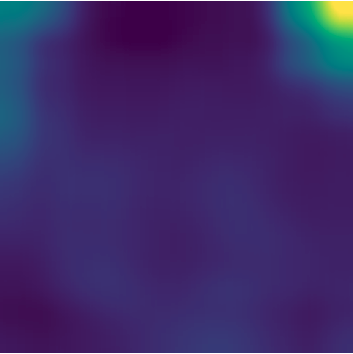}} &
  {\includegraphics[width=0.180\linewidth]{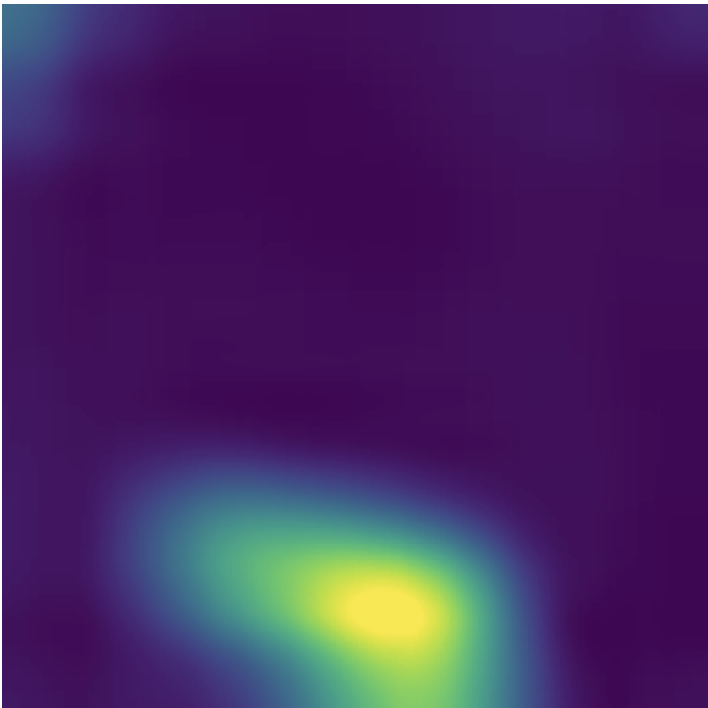}}&
  {\includegraphics[width=0.180\linewidth]{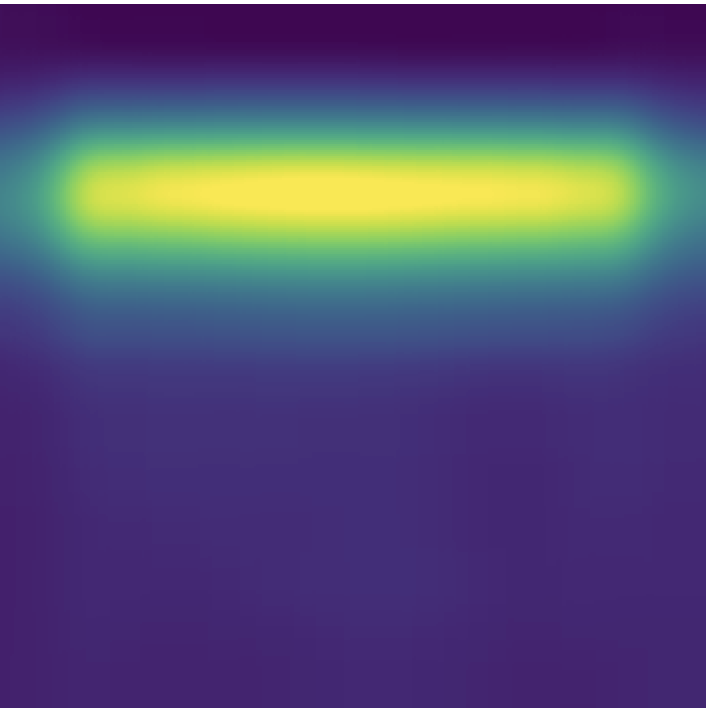}}&
  {\includegraphics[width=0.180\linewidth]{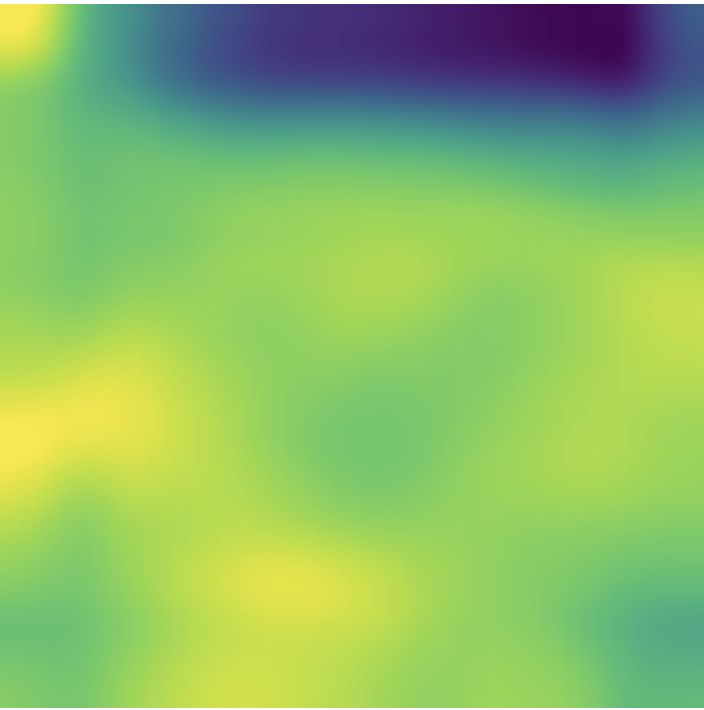}} &
  {\includegraphics[width=0.180\linewidth]{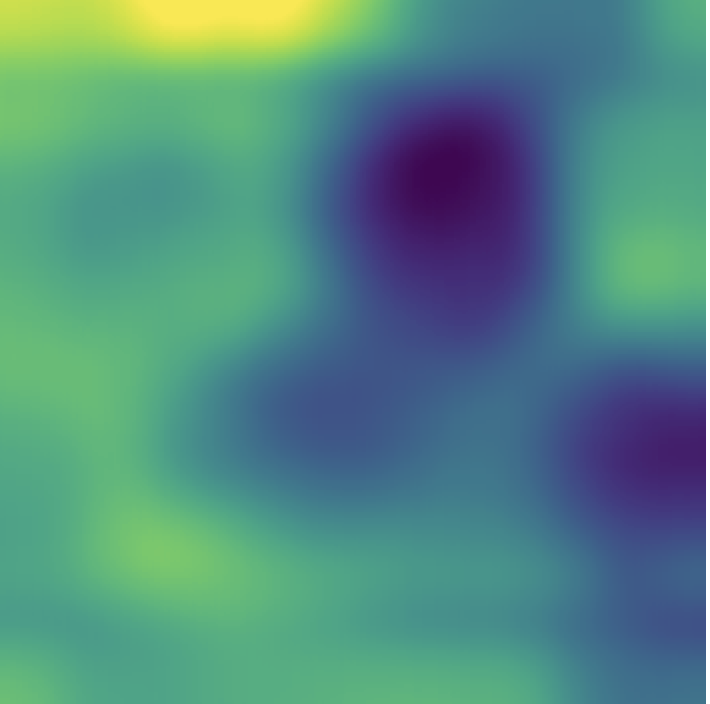}}\\
\footnotesize{(a)} & \footnotesize{(b)} & \footnotesize{(c)}& \footnotesize{(d)}& \footnotesize{(e)}\\
  \end{tabular}
  \end{center}
  \vspace{-6mm}
\caption{\small Observed confounding concepts in \textit{ISIC2019-2020 }datasets, the top row shows sample images, and the bottom row is the corresponding heatmap from GradCAM: (a) dark corners. (b) rulers. (c) dark borders. (e) dense hairs. (f) air pockets.}
  \label{figure2}
  \vspace{-4mm}
\end{figure}

To the end, we propose a human-in-the-loop framework that can effectively discover and remove the confounding behaviors of the classifier by transforming its feature representation into explainable concept scores. Then, humans are allowed to intervene based on first-order logic. Through human interaction (see Fig.~\ref{fig1}) on the model's CAVs, people can directly provide feedback to the model training (feedback turned into gradients) and remove unwanted bias and confounding behaviors. Moreover, we notice that no suitable dermatology datasets are available for confounding behavior analysis. To increase the methods' reproducibility and data accessibility in this field, we curated a well-controlled dataset called \textit{ConfDerm} containing 3576 images based on well-known \textit{ISIC2019} and \textit{ISIC2020} datasets. Within the training set, all images in one of the classes are confounded by one of five confounding factors, including dark corners, borders, rulers, air pockets, and hairs. Still, in the testing set, all images are random. Some confounding factors within the dataset are illustrated in Fig. \ref{figure2}.

We summarize our main contributions as: 
(1) We have crafted a novel dataset called ConfDerm, which is the first skin lesion dataset used for systematically evaluating the model's trustworthiness under different confounding behaviors within the training set.
% (1) We have crafted a new dataset called \textit{ConfDerm}that is used for systematically evaluating the model's trustworthiness under different confounding behaviors within the training set.
(2) Our new spectral relevance analysis algorithm on the most popular skin cancer dataset \textit{ISIC2019-2020} has revealed insights that artifacts and biases such as dark corners, rulers, and hairs can significantly confound modern deep neural networks.
(3)  We developed a human-in-the-loop framework using concept mapping and logic layer rewriting to make the model self-explainable and enable users effectively remove the model's confounding behaviors.
% (3) We developed a human-in-the-loop pipeline using xxx to make the model self-explainable and enable user effectively remove the model's confounding behaviors.
% (4) We propose a new interaction method to make human involves in the training loop using first-order logic.
(4) Experimental results on different datasets demonstrate the superiority of our method on performance improvement, artifact removal, skin tone debiasing, and robustness on different testing sets.

\begin{figure*}[!t]
   \begin{center}
   {\includegraphics[width=0.93\linewidth]{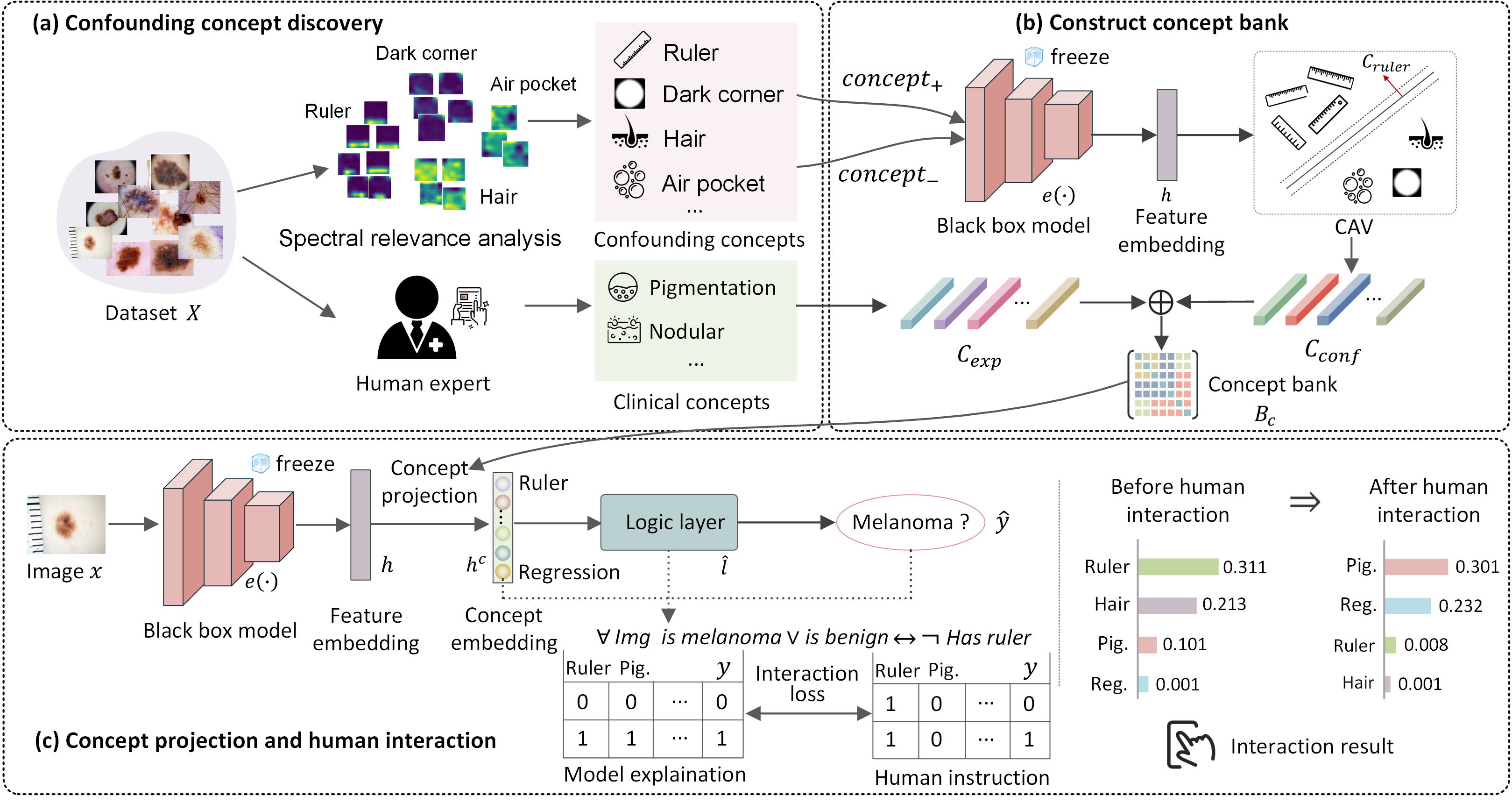}}
   \end{center}
%   \vspace{-10mm}
%   \vspace{-6mm}
\caption{\small 
Illustration of our human-in-the-loop pipeline. (a) Applying the improved spectral relevance analysis algorithm to discover the confounding factors within the dataset (see subsection \ref{sec:Global Confounding Concepts Discover}). (b) Learning confounding concept and clinical concept vectors (see subsection \ref{sec: Construction of Concept Bank}. (c) Projecting feature representation of a model onto concept subspace and then removing the model's confounding behaviors via human interaction (see subsection \ref{sec:Model Logic Rewriting}). Pig. denotes pigmentation, and Reg. denotes regression structure.
}
   \label{fig:overview}
   
\end{figure*}
\section{Related Work}
\subsection{Trustable Skin Diseases Diagnosis}
 In dermatology, clinicians usually diagnose skin cancer by assessing skin lesions based on criteria such as the ABCD rule \cite{abcd} and the seven-point checklist \cite{7pt}. As visual examination for skin cancer diagnosis is subjective and relies on the experience of dermatologists, deep learning-based diagnostic algorithms have been gaining attraction. Recently, Deep Learning methods \ie, DeepDerm \cite{deepderm}, have reached dermatologist-level diagnostic accuracy. However, the black-box nature of deep learning methods in the decision-making process has been a big hurdle to making clinicians trust them. Several researchers try to make skin cancer diagnosis trustable based on either attribution-based explanation methods (\ie, Integrated Gradients \cite{ig}, GradCAM \cite{gradcam}, LIME \cite{lime}) or concept-based explanation methods (\ie, Concept bottleneck Model \cite{CBM}, Concept Activation Vector \cite{cav}, and Posthoc Concept Model \cite{pcbm}). Our work follows the concept-based explanations but is different from them in the following aspects:
(1) Existing concept-based models for skin disease diagnosis only consider clinically related concepts and do not have the ability to find spurious correlations within the training data, making them can not conduct fine-grained error analysis.
(2)  we further studied using interactive learning to remove confounding behavior, which is an important but rarely explored point in skin cancer diagnosis and is also an essential step in building trustworthiness between clinicians and models.
\subsection{Concept-based Explanations}
Concept-based explanations identify high-level semantic concepts that apply to the entire dataset that is not limited by the feature space of deep neural networks. It can be roughly divided into two categories, concept bottleneck network (CBM) like methods \cite{CBM,CTransformer,logic_CBM}, and Concept Activation Vectors(CAVs) like methods \cite{cav,pcbm,skin_cav}. The CBM, as an interpretability-by-design model, first predicts the concepts, then predicts the target labels using the concepts by restricting the model's architecture. We argue that this method is not suitable for skin cancer diagnosis since CBM assumes the relevance between concepts and tasks are known in advance, but it is not the case when we want to solve new tasks. Also, the per-image level concept annotation is very expensive, especially for skin cancer datasets requiring expert annotation. Different from CBM, the CAVs train linear classifiers to the neural network's feature representation to verify whether the representation can separate the concept examples defined by people. Then, the coefficients of the linear classifier are CAVs. With the CAVs, we can get global explanations for a model's overall behaviors. As the user-defined concept examples are relatively easily obtained for different tasks (only need 20-70 images for each concept as the exemplar), we make use of CAVs in our work.

\subsection{Explanatory Interactive Learning}
Explanatory Interactive Learning (XIL) is a combination setting of Explainable AI (XAI) and active learning. It incorporates the human user in the training loop by querying the model's explanations and then giving feedback to correct the model's explanations. Some works \cite{human_attention,interactive_attention} tend to force the model to attend to important regions in the input images via attention learning, but attention is often different from explanation \cite{human_attention,no_exp}. Another line is to constrain the model's explanation by penalizing the model's gradient outside the correct regions to align human knowledge. \cite{inter_align,xil_nature} achieve it by providing the model's human-annotated masks during training. However, it is too expensive to pixel-wise annotate every image in skin lesion datasets.  \cite{Nesy} tends to compile a lot of concept labels for simulated datasets and changes the model's behavior via intervention on its concept representation of the model. Unfortunately, this method can not be applied to skin cancer diagnosis as it requires us to know and annotate all possible concepts in our task; this is not possible for real-world cases. 

%-------------------------------------------------------------------------

\section{Method}
To formalize our problem, consider a training set $\mathcal{D}=\lbrace (x_{i},y_{i})\rbrace_{i=1}^{N} $ where $x_{i}$ is an input image, $y_{i}$ is the corresponding ground truth, and N is the size of the training dataset, and a CNN classification model $ f= l \circ e$ where the feature extractor $e$ maps $x_{i}$ into latent embedding $h_{i}$ and then the classifier $l$ maps the $h_{i}$ into final prediction $y_{i}=l(h_i)$. Our objective is to identify the confounding factors $C_{conf}=\lbrace(c_{j})\rbrace_{j=1}^M$ within the training set $\mathcal{D}$ that can confound $f$, and then remove them via XIL on the explainable version of $f$. 

The overall pipeline of our method is summarized in Fig.~\ref{fig:overview}. First,  we apply spectral clustering \cite{spectral} on the GradCAM \cite{gradcam} computed with $f$ to discover different confounding concepts within the dataset $D$ (see Fig.~\ref{fig:overview} (a) and detail in section \ref{sec:Global Confounding Concepts Discover}). Then, we construct a concept bank $B_{c}$, which is composed of confounding concepts $C_{conf}$ and clinical-related expert concepts $C_{exp}$. We learn concept activation vectors (CAVs) from different clusters to obtain $C_{conf}$, and learn CAVs from a small probe dataset with expert-level concept annotations to obtain $C_{exp}$ (see Fig.~\ref{fig:overview} (b) and section \ref{sec: Construction of Concept Bank}). Next, we project the feature representation $h$ from extractor $e$ onto a concept subspace $h^{c}$ spanned by $B_{c}$. We replace the $l$ with an explainable logic layer $\hat{l}$ to model the relationship between concepts and the final prediction $\hat{y}=\hat{l}(h)$ based on the concept representation $h^{c}$. Finally, we rewrite the model's decision during training by applying interaction loss on the input gradient of $\hat{l}$. The details are elaborated in Fig.~\ref{fig:overview} (c) and section \ref{sec:Model Logic Rewriting}.

\subsection{Global Confounding Concepts Discover}
\label{sec:Global Confounding Concepts Discover}
In this section, we introduce our global confounding concept discovery algorithm (GCCD), which is used to semi-automatically discover the model's common confounding behaviors within the training set. Our method is based on spectral relevance analysis (SpRAy) \cite{spay}, originally used to analyze the co-occurring behavior with a dataset. 
As illustrated in Algorithm \ref{alg1}, given a training set $X$, class labels $C$, and a model $f$ trained on the $X$ \footnote{in practice, we can also directly choose a model trained on ImageNet.}, we want to visualize and obtain concept clusters in a 2D embedding $E$. 
First, we randomly sample a subset $\hat{X}$ from X. Then, we calculate the GradCAM heatmaps of images from $\hat{X}$ for each class. Instead of only performing prepossessing and normalization on heatmaps like SpRAy, we additionally apply discrete Fourier Transformation \cite{fourier} to distinguish different models' behaviors better to obtain high-quality concept clusters for similar but different concepts. Then, we concatenate each heatmap with its corresponding image to provide additional appearance information, which is different from SpRAy, which highly relies on the location information of the heatmaps. To accelerate the clustering process, we downscale the concatenated images five times, as suggested in \cite{spay}. Similar to \cite{rrr_nmi,spay}, we choose spectral clustering \cite{spectral} to calculate a spectral embedding based on an adjacency matrix calculated by the K-nearest neighborhood algorithm (KNN). To analyze the clusters, we perform non-linear dimension reduction via t-SNE \cite{tsne}. Finally, we manually annotate the concept label for each cluster and filter out those non-representative clusters.

% In the SpRAy, the author only uses GradCAM heat maps, which causes the clustering algorithm highly relies on the location information from heatmaps. Alternatively, we choose to concatenate each heatmap with its corresponding image, which can provide additional appearance information. 
\begin{algorithm}[!htb] 
\footnotesize
\caption{ Global Confounding Concept Discovery (GCCD).} 
\label{alg1} 
        \KwIn{Training set $X=\lbrace x_{i}\rbrace_{i=1}^{N} $\\
              \Indp \Indp
              \hskip-0.1em Class labels $C=\lbrace c_{j} \rbrace_{i=1}^{M}$\\
              Model $f$  \\ 
          }
          \KwOut{t-SNE embedding $E=\lbrace e_{k}\rbrace_{k=1}^{M} $}
          
Randomly sample a subset of X as $\hat{X}$\; 
    \For{$C_{j} \in C$} 
    { 
        $M_{cj}=\lbrace \rbrace$\; 
          \For{$\hat{x_{i}} \in \hat{X}$} {
              $M_{j}=GradCAM(f,\hat{x}_{i},c_{j})$\;
              $M_{j}=FourierTransform(M_{i})$\;
              $\hat{x_{i}}=Preprocessing(\hat{x_{i}})$\;
              $M_{j}=M_{j} \oplus \hat{x_{i}}$\;
              $M_{j}=Downscale(M_{j})$\;
              $M_{cj} \leftarrow M_{j}$\;
          }
      
        Obtain adjacency matrix $H_{j}=KNN(M_{cj})$\;
        Spectral clustering on $H_{j}$, get the sepctral embedding $\phi_{j}$\;
        Visualize 2D embedding $e_{k}=tSNE(\frac{1}{H_{j}+\epsilon})$\;
    } 
    return $E$
\end{algorithm}

\subsection{Construction of Concept Bank}
\label{sec: Construction of Concept Bank}
To build our concept bank $C_{conf}$, we make use of Concept Activation Vectors (CAVs) \cite{cav}. For each concept, given a series of positive concept examples $P^{c}=\lbrace P_{i}^{c}\rbrace_{i=1}^{T_{P}}$ and negative concept examples $N^{c}=\lbrace N^{c}_{i}\rbrace_{i=1}^{T_{N}}$. We train a linear classifier to separate the CNN features of those examples that contain the concept $e(P^{c})$ or not $e(N^{c})$. Then, the CAVs are defined as a normal $w^{c}$ to a hyperplane that separating $e(P^{c})$ from $e(N^{c})$ in the embedding space $h$, such that satisfying $(w^{c})^{T}h+\phi^{c}>0$ for all $h \in e(P^{c})$ and $(w^{c})^{T}h+\phi^{c}<0$ for all $h \in e(N^{c})$ where $w^{c}$ and $\phi^{c}$ is the weight and bias of the linear classifier. 

For our case, we collect confounding concepts $C_{conf}$ from previously generated concept clusters $E$ %generated with the method mentioned in Section \ref{sec:Global Confounding Concepts Discover},
with a human confirmation. To collect clinical-related concepts $C_{exp}$, we collect concepts from existing expert-level annotated probe datasets (\ie{12 concepts from \textit{Derm7pt}\cite{derm7pt} for dermatoscopic images or 22 concepts from \textit{skincon}\cite{skincon} for clinical images}). Finally, we get the concept bank $B_{c}=C_{conf} \oplus C_{exp}$; more details will be described in the Experimental part.

\subsection{Model Logic Rewriting}
\label{sec:Model Logic Rewriting}
    \noindent\textbf{Turn the black box into explainable models:}
    \label{concept mapping}
    A black-box model $f$ (could be any backbone \ie, ResNet, Inception, or  Vision transformer) can be simplified as a composition of a feature extractor $e$, and classification decision layers $l$. Our aim is to make it logically interpretable to increase its trustworthiness. First, we project hidden embedding features $h$ of $e(X)$ onto a low dimensional concept embedding $h^{c}$. Each dimension of $h^{c}$ corresponds to a concept in $B_{c}$. Specifically, we define $h^{c}=\frac{\langle h \; , B_{c}\; \rangle }{\lVert B_{c} \rVert^{2}}B_{c}$ where $h^{c}$ is concept scores, $B_c$ is the concept bank containing CAVs for all concepts. 
    We replace the $l$ with an explainable layer $\hat{l}$, so that $ \hat{l}(h^{c})=\hat{y} $ where its prediction $\hat{y}$ is based on the explainable concept embedding $h^{c}$. The linear layer, decision tree, or logic layer can be used as the explainable layer $\hat{l}$. In the medical community, dermatologists diagnose a lesion based on combining prediction results for different clinical concepts (\ie{ABCD rule \cite{abcd} and seven-point checklist \cite{7pt}}) to get a diagnosis score. A lesion is diagnosed as melanoma if its diagnosis score is larger than a threshold. To mimic the skin cancer diagnosis process and produce human understandable interpretability, we choose the recently proposed entropy-based logical layer \cite{logic_CBM} as it can binarize the concepts and model their relationship based on importance weights to perform final decision. More specifically, it produces first-order logic-based explanations by performing an attention operation on the model's concept representation. However,\cite{human_attention,no_exp} show that attention is often not the explanation, which causes interaction on it is not effective in changing the model's behavior (see examples and module details in supplementary material). \\
    \noindent\textbf{Interaction and model re-writing:} To generate faithful explanations of the logic layer $\hat{l}$, we employ input gradient explanation to highlight the important concepts in the concept representation $h^{c}$. Given $h^{c} \in b \times d$ and $\hat{l} \in b \times d \rightarrow b \times p$ where $b$ is the number of samples, $d$ is the number of concepts, and $p$ is the number of classes, the input gradient is defined as $\nabla_{h^{c}} \hat{l}(h^{c})$, which is a vector norm to the model's decision boundary at $h^{c}_{i}$, and can be regarded as a first-order description of model's behavior around $h^{c}_{i}$. The users are able to intervene in the model's behavior by providing an annotation matrix $H \in b \times d$ to model $\hat{l}$, where $H$ represents the first-order logic used to inform the model which concepts are irrelevant to its prediction. We achieve this by employing the right for the right reason (RRR) loss \cite{rrr}. We replace the $L2$ regularization term with the elastic-net regularizer to improve the model's sparsity, and therefore users can more easily intervene with its explanation. Our final loss is defined as: %With the general cross-entropy classification loss, RRR loss, and elastic-net regularization,

 \begin{equation}
 \vspace{4mm}
 \label{eq1}
 \footnotesize
    \begin{split}
      \mathcal{L}(h^{c}, y, H )&=\underbrace{\mathcal{L}_{CE}(y,\hat{y})}_\text{Cross Entropy Loss}+\underbrace{\lambda_{1}(\alpha \Vert w \Vert_{1} + (1-\alpha) \Vert w \Vert_{2}^{2}  ) }_\text{Elastic-Net Regular}] \\ & + \underbrace{\lambda_{2} \sum_{b=1}^{B}\sum_{d=1}^{D}(H_{bd}\frac{\partial}{\partial h_{bd}^c}\sum_{c=1}^{N^{c}}(\hat{y}_{bc}))^2}_\text{Right Reasons Loss}\\
    \end{split}
 \end{equation}
where the right reason loss penalizes the input gradient from being large in the positions annotated by a human user (see the graph at the bottom of Fig.~\ref{fig:overview}). The $\lambda_{1}$ controls the sparsity of $\hat{l}$, and $\lambda_{2}$ controls how much the human user instruction $H$ is considered by the model.

 \begin{figure}[!t]
\setlength{\abovecaptionskip}{0.cm}
\setlength{\belowcaptionskip}{-0.cm}
   \begin{center}
   {\includegraphics[width=0.95\linewidth]{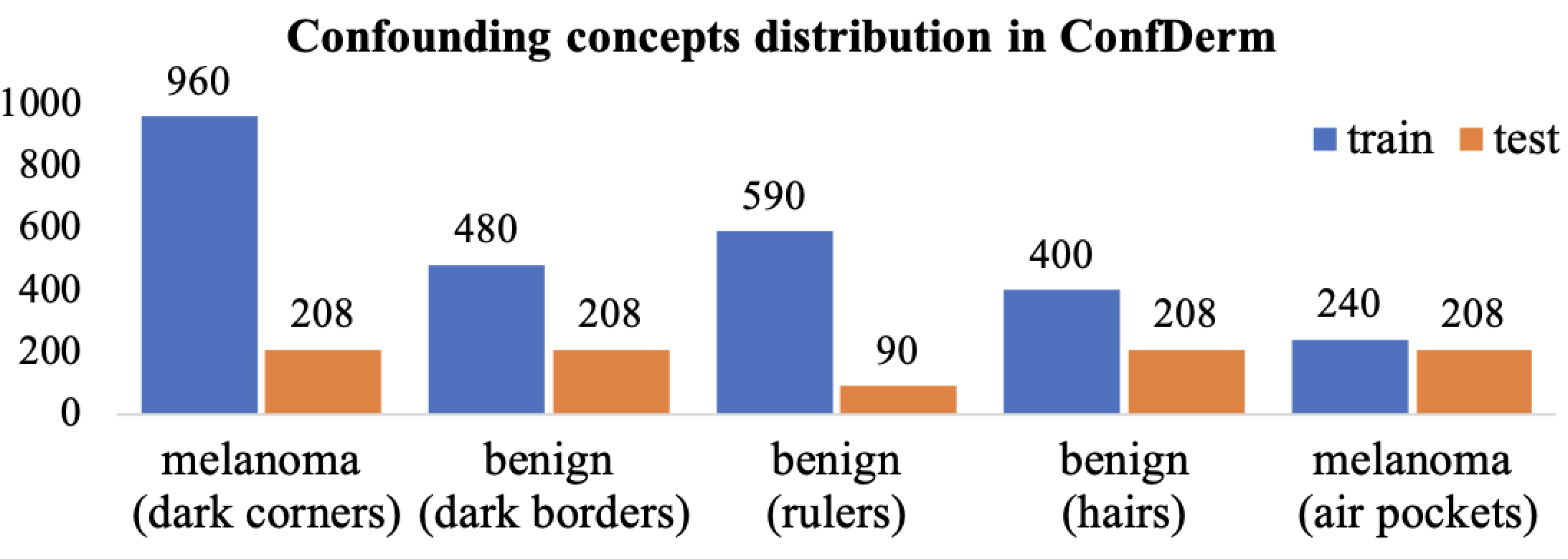}}
   \end{center}
   \vspace{-2mm}
\caption{\small Confounding concepts distribution in our proposed \textit{ConfDerm} dataset.}
\vspace{-4mm}
   \label{fig:figure6}
\end{figure}
\section{ConfDerm Dataset}
 We found it difficult to evaluate the model's confounding behavior on existing dermatology datasets due to two reasons: (1) removing confounding behaviors within a dataset does not always improve performance, especially when the distribution in the testing set is similar to the training set (\ie, both have dark corners). (2) It is also hard to quantify all confounding factors in a real skin lesion dataset. To alleviate these problems, we collect a well-controlled dataset \textit{ConfDerm} with 3576 real dermoscopic images with melanoma and benign classes from \textit{ISIC2019}\cite{{isic2019}}, and \textit{ISIC2020}\cite{isic2020}. Our dataset consists of five sub-datasets to capture different confounding factors for a particular class, including \textit{melanoma (dark corners)}, \textit{benign (dark borders)}, \textit{benign (rulers)}, \textit{benign (hairs)}, and \textit{melanoma (air pockets)}. For example, in the \textit{benign (dark borders)} dataset, all benign images contain dark borders in the training set but no dark borders in the testing set. Still, all benign images in the training and testing set are random. The concept distribution of all five datasets is described in Fig.~\ref{fig:figure6}, and we omit the class distribution of each sub-dataset as classes in all datasets are balanced.

\section{Experiments}
\begin{figure*}[!t]
   \begin{center}
  {\includegraphics[width=0.98\linewidth]{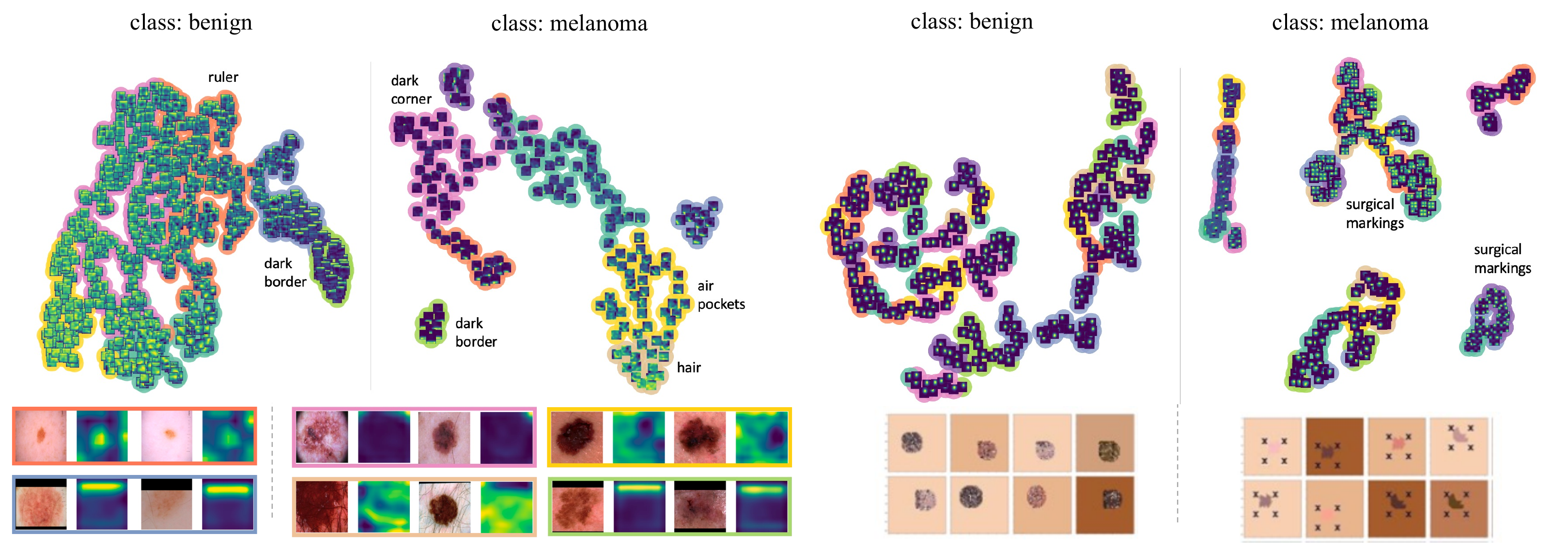}}
 
   \end{center}
  \vspace{-4mm}
\caption{\small Global analysis of the models' behavior within datasets using GCCD. The left graph is the tSNE of spectral clustering using GradCAMs of a ResNet50 within \textit{ISIC2019} and \textit{ISIC2020}. The right one is the tSNE of spectral clustering using GradCAMs of a VGG16 within the \textit{SynthDerm}dataset.}
   \label{fig:fig7}
\end{figure*}

\subsection{Experimental setup}

Our task is to classify a skin lesion into benign or melanoma. 
To evaluate the trustworthiness of the above task of using conventional CNN vs. our proposed method, we perform \ {three} main experiments on five public training sets and one novel dataset we crafted: \textbf{(1) confounding concepts discovery}: we use our GCCD (see Algorithm.\ref{alg1}) to discover the confounding factors on three popular skin lesion datasets and one synthetic dataset. \textbf{(2) rewriting model's decision:} we perform our human interaction method on our novel dataset \textit{ConfDerm} to evaluate its effectiveness in handling five challenging confounding problems of dermoscopic images.  \textbf{(3) Debiasing the Effect of Skin Tone:} we also study the effect of skin tone using our interaction method. We evaluate model robustness across datasets with different skin tone scales by removing the skin tone concept.\\\
\noindent\textbf{Training setting:} 
Our method is general and can be applied to any backbone, such as VGG \cite{vgg}, ResNet \cite{resnet}, Inception \cite{inception}, and ViT \cite{vit}. For subsection \ref{sec_exp1} and \ref{sec_exp2}, we use ResNet50 as the backbone model, and for subsection \ref{sec_exp3}, following the setting in \cite{deepderm,ddi},  we use Inception V3 with DeepDerm weights as our backbone and fine-tune it on \textit{Fitzpatrick 17K} dataset. For evaluation metrics, we use accuracy for experiments in subsection \ref{sec_exp2} as the dataset \textit{ConfDerm} has balanced classes, and we use ROC-AUC for all other skin cancer diagnosis experiments. For hyper-parameters, prepossessing and more detail of each used dataset is in the supplementary material.  \\
\noindent\textbf{Human interaction:} Most existing XIL settings  \cite{cd,rrr_nmi} either use per-sample level human-annotated masks or concept labels to simulate the human user to have full knowledge about the task. However, it is often not realistic and general for skin cancer diagnosis as most skin lesion datasets do not have these annotations, and these annotations are also very expensive. Different from these local interaction methods, we choose to use global interaction similar to \cite{Nesy} on confounding factors. For example, we provide a rule table defined by human-user in the column of the concept ``rulers" is all 1s, and other columns are all 0s, which means ``never focus on rulers" (see Fig. \ref{fig:overview} (c)). This is an efficient method to perform interaction as we only need one rule table to change the model's behavior for the entire dataset. More details about interaction are described in the following subsections.
\subsection{Confounding Concept Discovery}
\label{sec_exp1}
We perform our GCDD algorithm (see subsection \ref{sec:Global Confounding Concepts Discover}) on four skin lesion datasets. Specifically, we train ResNet50 on \textit{ISIC2016} \cite{isic2016}, \textit{2017} \cite{isic2017}, and \textit{2019-2020} \cite{isic2019,isic2020} and train VGG16 on the \textit{SynthDerm} \cite{synthderm} dataset where it contains a confirmed confounding factors ``surgical markings".
% We perform our GCDD algorithm (see subsection \ref{sec:Global Confounding Concepts Discover}) on four skin lesion datasets. Specifically, we train ResNet50 on \textit{ISIC2016 }\footnote{https://challenge.isic-archive.com/landing/2016/}, \textit{2017}\footnote{https://challenge.isic-archive.com/landing/2016/}, and \textit{2019-2020}\footnote{https://www.kaggle.com/datasets/qikangdeng/isic-2019-and-2020-melanoma-dataset} and train VGG16 on the SyntheDerm\footnote{https://affect.media.mit.edu/dissect/synthderm/} dataset where it contains a confirmed confounding factors ``surgical markings". 
We summarized the discovered confounding concepts from all four datasets in Table \ref{tab1:confdiscover}.
The ResNet50 trained on \textit{ISIC2019-2020 }can achieve 86.36 \% ROC-AUC on the \textit{ISIC2020} testing set, and VGG16 can achieve 100\% accuracy on SynthDerm. However, by visualizing the clusters discovered by GCDD on \textit{ISIC2019-2020 }(see Fig. \ref{fig:fig7} left), it shows that ResNet50 predicts melanoma often based on confounding concepts, including dark corners, dark borders, hairs, and air pockets, while it predicts benign often based on dark borders or rulers. As for GCDD on SynthDerm, in the right of Fig. \ref{fig:fig7}, it shows that VGG16 highly relies on surgical markings when predicting melanoma.  
More experimental results and visualization are provided in the supplementary material.
\begin{table}[t!]
  \centering
  \scriptsize
  \renewcommand{\arraystretch}{1.2}
  \renewcommand{\tabcolsep}{2.8mm}
  \caption{The discovered confounding concepts from different skin lesion datasets. DC denotes dark corners, DB denotes dark borders, RL denotes rulers, APs denotes air pockets, and SMs denotes surgical markings.
  }
    \vspace{-2mm}
  \begin{tabular}{c|cccccc}
  \hline
%   \toprule
    Skin Lesion Dataset &DC&
  DB  & RL  & HR & APs    & SMs  \\
  \hline
%   \multirow
    \textit{ISIC2019-2020 }   & \checkmark& \checkmark & \checkmark & \checkmark& \checkmark & \\
    \textit{ISIC2016 }   & \checkmark& \checkmark &  & & &  \\
    \textit{ISIC2017}       & \checkmark& \checkmark & &\checkmark&  &  \\
    \textit{SynthDerm}&& & && & \checkmark\\
    \hline

  \hline
  \end{tabular}
  \vspace{-2mm}
  \label{tab1:confdiscover}
\end{table}
% \begin{table}[t!]
%   \centering
%   \scriptsize
%   \renewcommand{\arraystretch}{1.3}
%   \renewcommand{\tabcolsep}{2.8mm}
%   \caption{The discovered confounding concepts from different skin lesion datasets. DC denotes dark corners, DB denotes dark borders, RL denotes rulers, APs denotes air pockets, and SMs denotes surgical markings.
%   }
%   \begin{tabular}{c|cccccc}
%   \hline
% %   \toprule
%     Skin Lesion Dataset &DC&
%   DB  & RL  & HR & APs    & SMs  \\
%   \hline
% %   \multirow
%     \ isicISIC\ \textit{ISIC2019-2020 }   & \checkmark& \checkmark & \checkmark & \checkmark& \checkmark & \\
%     \ isicISIC\ \textit{ISIC2016 }   & \checkmark& \checkmark &  & & &  \\
%     \ isicISIC\ \textit{ISIC2017}   & \checkmark& \checkmark & &\checkmark&  &  \\
%     \textit{SynthDerm}&& & && & \checkmark\\
%     \hline

%   \hline
%   \end{tabular}
%   \vspace{-2mm}
%   \label{tab1:confdiscover}
% \end{table}

 \begin{figure}[!t]
\setlength{\abovecaptionskip}{0.cm}
\setlength{\belowcaptionskip}{-0.cm}
   \begin{center}
   {\includegraphics[width=0.99\linewidth]{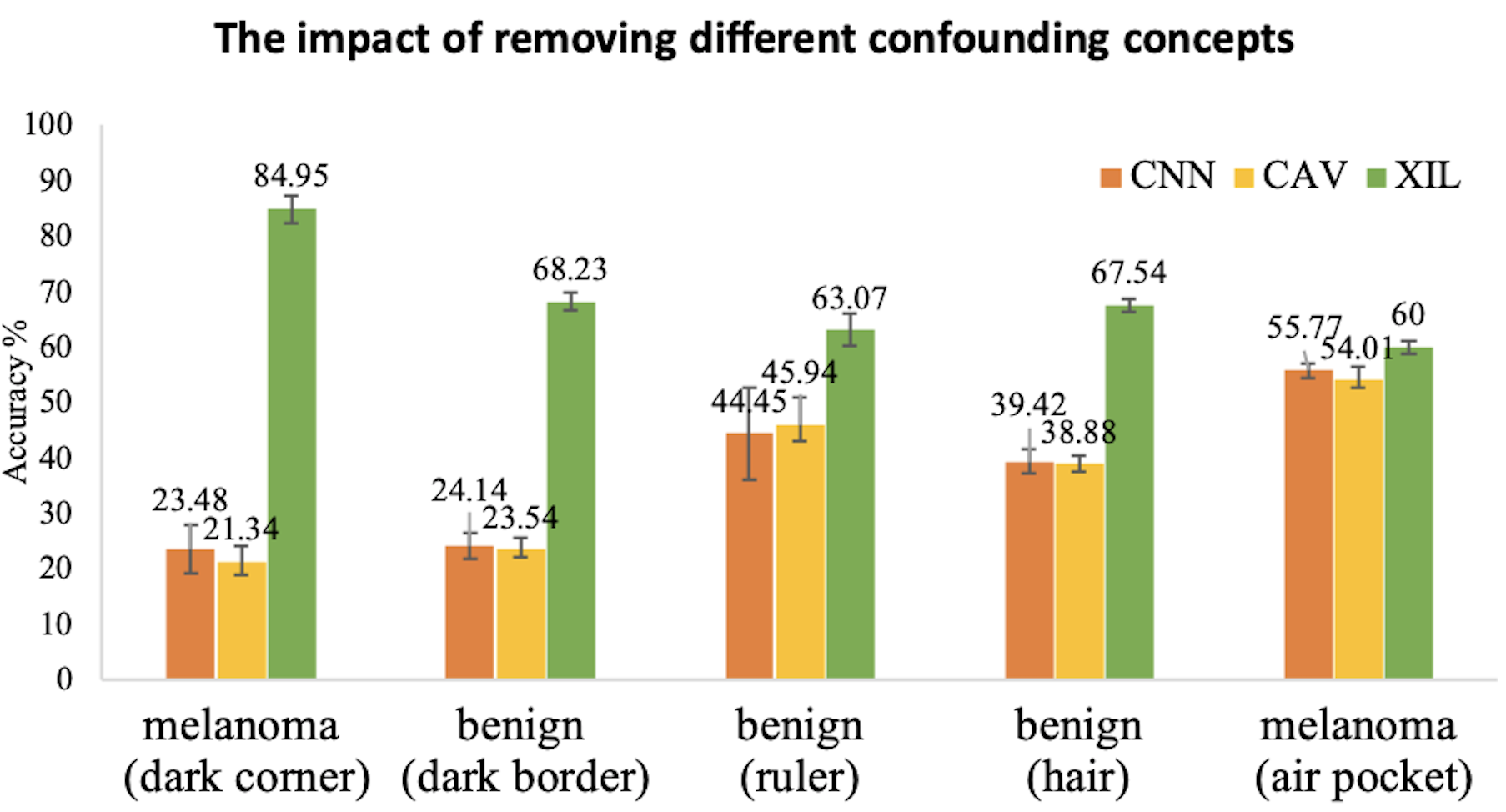}}
   \end{center}
   \vspace{-2mm}
\caption{\small Performance improvement on confounded class when removing different confounded classes using XIL in five confounded datasets in the \textit{ConfDerm} dataset.}
\vspace{-4mm}
   \label{fig:figure8}
\end{figure}

\subsection{Rewriting Model's Decision in ConfDerm}
\label{sec_exp2}

\begin{figure*}[!t]
   \begin{center}
      \begin{tabular}{ c@{ } c@{ } }
  {\includegraphics[width=0.48\linewidth]{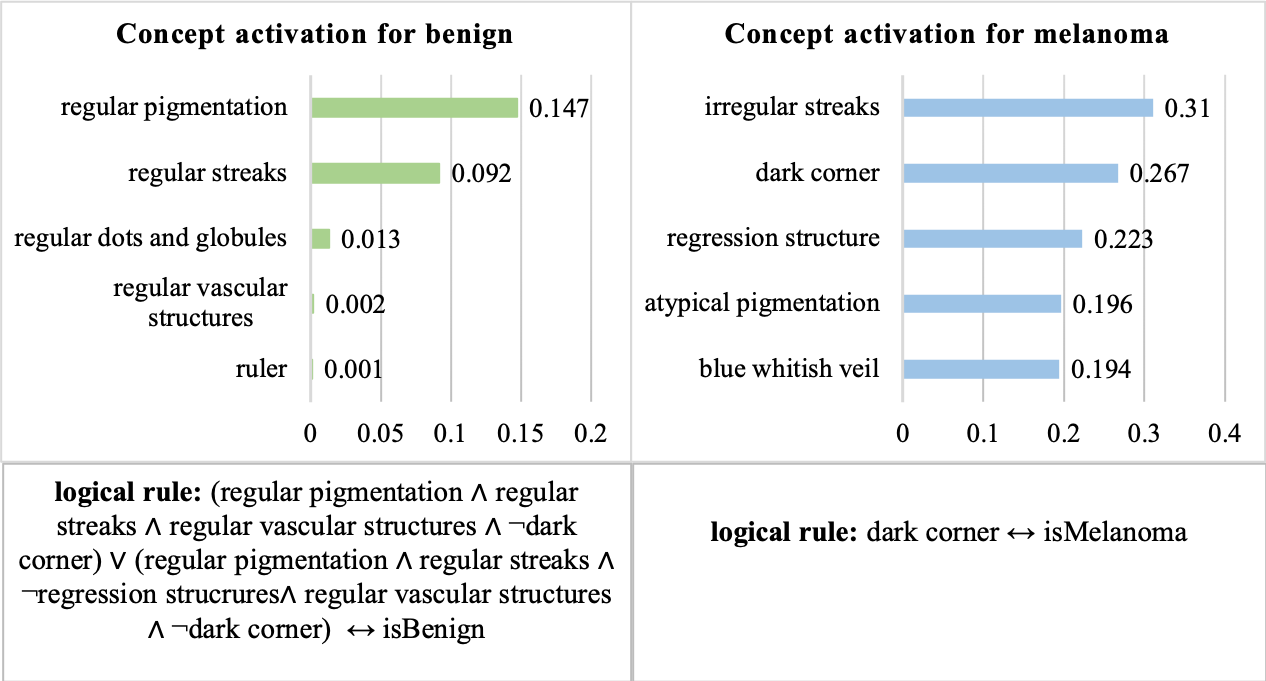}}&
  {\includegraphics[width=0.48\linewidth]{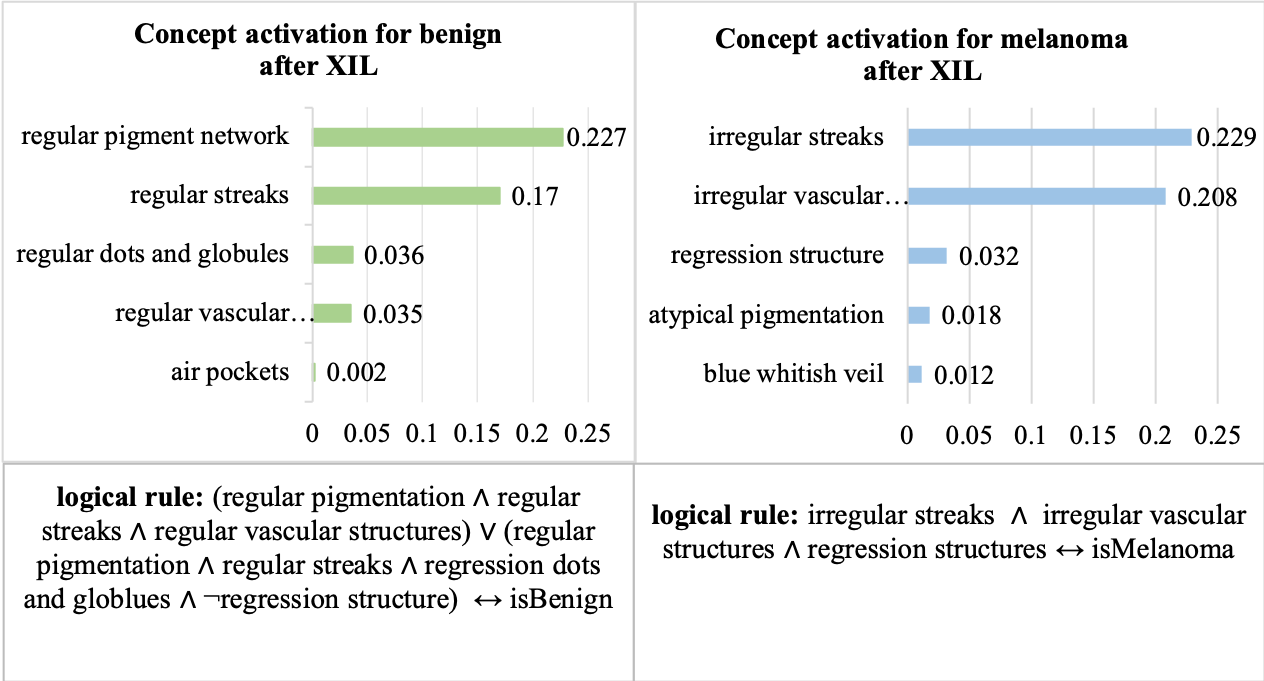}}
    \end{tabular}
   \end{center}
      \vspace{-4mm}
%   \vspace{-10mm}
\caption{\small The global explanation of the model's behavior on the \textit{melanoma (dark corners)} dataset of ConfDerm. In the left two figures, either the concept activation or logical rule shows that the model is confounded by the concept of the dark corners when predicting melanoma. In the right two figures, after the interaction, the model does not predict melanoma based on the dark corners, and it predicts melanoma based on meaningful clinical concepts (Zoom for better visualization).}
   \label{fig:fig9}
\end{figure*}

In this section, we perform experiments on our designed \textit{ConfDerm} dataset, based on \textit{ISIC2019-2020}, to verify whether our interaction (XIL) method can remove the confounding concept for each sub-dataset. We collect 12 clinical concepts from \textit{Derm7pt}\cite{derm7pt} probe dataset and five confounding concepts from the clusters produced on \textit{ISIC2019-2020 }(see Fig.~\ref{fig:fig7} left). For each concept, we collect 70 positive samples and 70 negative samples to learn CAVs using a linear SVM. We define one global rule for each sub-dataset to ignore a specific confounded concept like ``Do not focus on the dark corners, dark borders, rulers, hairs, and air pockets" for ``\textit{melanoma (dark corners)}", ``\textit{\textit{benign (dark borders)}}", ``\textit{benign (rulers)}", ``\textit{benign (hairs)}" and ``\textit{melanoma (air pockets)} sub-datasets respectively." We compare our interaction method ``XIL" with the CNN baseline and CNN with concept mapping (see ``Turn any black box into explainable models" in section \ref{concept mapping} ). The CNN baseline is a ResNet50 model, CAV is the CNN baseline with concept mapping, and XIL is CAV with our logical layer using human interaction learning. The impact of removing each confounding concept on the confounded class is shown in Fig.~\ref{fig:figure8}. For CAV, it shows that it can achieve comparable performance with CNN but cannot make the model robust to different confounding concepts. For our XIL method, the consistently better performance against CNN and CAV across all five sub-datasets demonstrates our XIL method's effectiveness in removing the model's confounding behavior. Also, it can be seen that XIL improves most on the \textit{melanoma (dark corners)} datasets; removing dark corners can improve the accuracy of CNN from 23.48\% to 84.95\% for the melanoma class. This inspires us dark corners may be the main artifacts to limit the accuracy of skin cancer diagnosis, and the quality of the skin lesion dataset can be improved by avoiding collecting images with dark corners. Also, XIL gets the minimum improvement on \textit{melanoma (air pockets)}; removing air pockets from melanoma only improves the accuracy from 55.77 \% to 60 \% for the melanoma class, and it may be the air pockets concept is relatively not a significant confounding factor for skin cancer diagnosis. Besides performance improvement, the correct decision-making behind the model is also essential for trustable diagnosis. In Fig.~\ref{fig:fig9}, both concept activation and logic explanation of our method show that the model does not perform its prediction based on the dark corners anymore after using interaction (we provide more examples in supplementary material). Moreover, we report the performance of the three methods for all classes on \textit{ConfDerm} with five random seeds in Table \ref{tab2}. It can be seen that XIL improves the performance for all classes on all five sub-datasets. These results demonstrate that human interaction can help models perform better on testing sets with different distributions. 
\begin{table}[t!]

  \centering
  \scriptsize
  \renewcommand{\arraystretch}{1.3}
  \renewcommand{\tabcolsep}{3.4mm}
  \caption{Performance on all five sub-datasets of \textit{ConfDerm}. MEL and BEN denote melanoma and benign. DC denotes dark corners, DB denotes dark borders, RL denotes rulers, HR denotes hairs, and APs denotes air pockets.}
    \vspace{-2mm}
  \begin{tabular}{c|ccc}
  \hline
  Datasets & CNN & CAV & XIL \\
  \hline
    MEL (DC) & $60.86 \pm 0.81$& $59.40 \pm 0.93$ &$\textbf{83.16} \pm \textbf{2.51}$\\

     BEN (DB) & $72.13 \pm 2.05$ & $73.40 \pm 1.63$& $\textbf{75.88} \pm \textbf{1.39} $\\

     BEN (RL) & $77.87 \pm 6.25$ & $77.93\pm 5.31 $& $\textbf{80.38} \pm \textbf{6.80}$ \\

     BEN(HR) & $64.33 \pm 2.11$ & $64.25\pm 2.74$ &$\textbf{74.91}\pm \textbf{1.52}$ \\

     MEL (APs) & $60.58 \pm 0.68$ & $60.44\pm 0.95$ &$\textbf{62.17} \pm \textbf{0.60}$\\

  \hline

  \end{tabular}
  \label{tab2}
\end{table}

\begin{figure}[t]
   \begin{centering}
      \begin{tabular}{ c@{ } c@{ } c@{ }}
  {\includegraphics[width=0.96\linewidth]{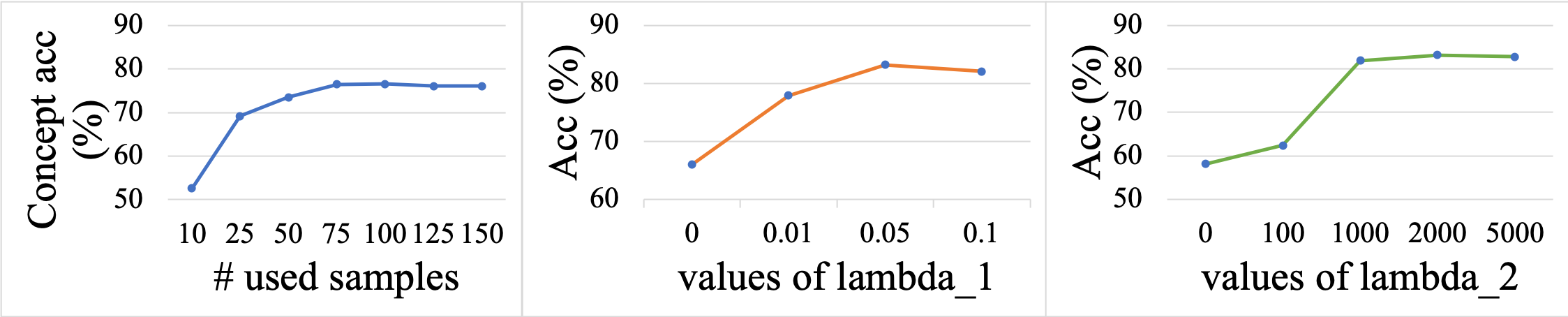}}
   \\
    \end{tabular}
   \end{centering}
      \vspace{-2mm}
%   \vspace{-10mm}
\caption{\small From left to right, the ablation on the number of used positive concept samples, the value of $\lambda_{1}$ of elastic-net regularization, and the value of $\lambda_{2}$ of the RRR loss.}
   \label{fig:ab}
\end{figure}
\subsection{Debiasing the Negative Impact of Skin Tone}
\label{sec_exp3}
As illustrated in \cite{ddi}, existing dermatology AI models perform worse on dark skin tones as dark skin images are under-represented in most popular dermatology datasets. In this experiment, we use our XIL method again to verify whether it can reduce the negative impact of skin tone. Specifically, we use the \textit{skincon}\cite{skincon} dataset as our probe dataset to collect 22 clinical concepts and one confounding concept ``dark skin". For each concept, we collect 50 positive samples and 50 negative samples and train each concept using a linear SVM. We fine-tune the DeepDerm model on the \textit{Fitzpatrick 17K} dataset (2168 images of dark skin types compared with 7755 images of the light skin type and 6098 images of the middle skin types) and then remove the dark skin concept using XIL. In Table \ref{tab3}, we compare our model with the DeepDerm without XIL on the \textit{DDI} dataset with Fitzpatrick skin tone scale labels. The consistently better performance of our method demonstrates that removing the dark skin tone concept can reduce the negative impact of skin tone and improve the robustness of the model across different skin tone type images. Further, our method achieve the most improvement on FST(\uppercase\expandafter{\romannumeral5}-\uppercase\expandafter{\romannumeral6}), which corresponds to dark skin tone images.  More results and visualizations are provided in the supplementary material.
\begin{table}[t!]

  \centering
  \scriptsize
  \renewcommand{\arraystretch}{1.4}
  \renewcommand{\tabcolsep}{1.3mm}
  \caption{Performance (ROC-AUC) improvement using XIL on dark skin concept on the \textit{DDI} dataset. FST (\uppercase\expandafter{\romannumeral1}-\uppercase\expandafter{\romannumeral2}) corresponds to light skin tones, FST (\uppercase\expandafter{\romannumeral3-\uppercase\expandafter{\romannumeral4}}) corresponds to middle skin tones, and \textit{DDI} is the combination of FST (\uppercase\expandafter{\romannumeral1-\uppercase\expandafter{\romannumeral6}}).}
\vspace{-2mm}
  \begin{tabular}{c|ccccc}
  \hline
  Method & DDI& FST (\uppercase\expandafter{\romannumeral1}-\uppercase\expandafter{\romannumeral2}) & FST (\uppercase\expandafter{\romannumeral3-\uppercase\expandafter{\romannumeral4}})&FST(\uppercase\expandafter{\romannumeral5}-\uppercase\expandafter{\romannumeral6})\\
  \hline
    CNN & $63.36 \pm 0.28$& $64.13 \pm 1.53$ &$68.43 \pm 0.70$&$57.933 \pm 1.47$\\

    CAV & $63.71 \pm 0.34$ & $64.34 \pm 1.93$& $67.92 \pm 0.73$ & $57.22 \pm 2.13$\\

    XIL & $\textbf{64.57} \pm \textbf{0.57}$ & $\textbf{65.32} \pm \textbf{0.54}$ & $\textbf{68.77} \pm \textbf{1.17}$ & $\textbf{60.03}\pm \textbf{1.79}$\\

  \hline

  \end{tabular}
  \label{tab3}
\end{table}
\begin{table}[t!]

  \centering
  \scriptsize
  \renewcommand{\arraystretch}{1.4}
  \renewcommand{\tabcolsep}{3.1mm}
  \caption{Comparison of the different interaction layers and interaction strategies (Accuracy for all classes).}
   \vspace{-2mm}
  \begin{tabular}{c|ccc}
  \hline
    & Editing w/ norm & L1 loss & RRR loss \\
  \hline
    Linear \cite{pcbm} & $57.21 \pm 0.03$& $54.30 \pm 0.47$& $47.09 \pm 0.39$\\

     MLP & \text{-} & $79.42 \pm 1.53$& $83.09 \pm 3.48$\\

     Logic Layer & $54.94 \pm 0.19$ & $79.17\pm 0.14 $& $\textbf{83.16} \pm \textbf{2.51}$ \\
  \hline
  \end{tabular}
  \label{tab4}
\end{table}
\subsection{Ablation Study}
\label{sec_ab}
We conduct experiments to analyze the impact of key hyper-parameters and components in our framework. For the impact of the used number of concepts, we analyze it on the \textit{Derm7pt} dataset. For other ablation studies, we perform experiments on the ``\textit{melanoma (dark corners)} dataset of ConfDerm, and we report the accuracy for all classes.  \\
\noindent\textbf{Impact of the used concept samples:} We perform experiments to analyze how many concept samples are suitable for learning a high-quality concept bank. We used 30 samples as the testing set and repeatedly ran the concept learning ten times with random seeds. The mean concept accuracy is shown on the left of Fig.~\ref{fig:ab}, and it can be seen that the performance saturates when using around 75 samples.\\
\noindent\textbf{Impact of $\lambda_{1}$ and $\lambda_{2}$:} The $\lambda_{1}$ and $\lambda_{2}$ in Eq. \ref{eq1} are two important hyper-parameters to control the contribution of sparsity and human interaction strength. By seeing the middle of Fig.~\ref{fig:ab}, we can see a relatively larger sparsity strength is helpful to improve performance as higher sparsity can make interaction easier. The right of Fig.~\ref{fig:ab} shows that setting a $\lambda_{2}$ with 1000 is enough for us to perform interaction to the model.\\
\noindent\textbf{Impact of different interaction layers:} In Table \ref{tab4}, we compare the performance when using different interaction layers, including the simple linear layer (posthoc-hoc CBM \cite{pcbm}), two-layer MLP, and the logic layer. We can see that two-layer MLP can achieve comparable performance with the logic layer, but the two-layer MLP cannot generate concept activations and is unable to model the relationship of concepts.\\
\noindent\textbf{Impact of different interaction strategies:} We compare different interaction strategies, including editing with normalization, L1 loss, and RRR loss in Table \ref{tab4}. It can be seen that editing with normalization is ineffective for the ``\textit{melanoma (dark corners)}" dataset, as editing with normalization will also hurt the performance of the unconfounded class (benign in this case). And we can see that L1 and RRR loss can alleviate this problem and achieve much better performance. The best performance of RRR loss demonstrates its effectiveness on the task.

\section{Conclusion}
In this paper, we show that confounding behaviors are common but rarely explored problems in skin cancer diagnosis. To tackle it, we propose a human-in-the-loop framework to effectively discover and remove the confounding behaviors of the model within the dataset during the skin cancer diagnosis. Moreover, to promote the development of skin cancer diagnosis, we crafted a new confounded dataset \textit{ConfDerm} with different distributions between the training and testing sets. Experimental results on different datasets demonstrate our method can improve the model's performance and trustworthiness in different testing distributions.

%------------------------------------------------------------------------

%%%%%%%%% REFERENCES
{\small

\bibliographystyle{ieee_fullname}
\bibliography{Trust_Derm_Arxiv}

\begin{thebibliography}{10}\itemsep=-1pt

\bibitem{7pt}
Giuseppe Argenziano, Gabriella Fabbrocini, Paolo Carli, Vincenzo De~Giorgi,
  Elena Sammarco, and Mario Delfino.
\newblock Epiluminescence microscopy for the diagnosis of doubtful melanocytic
  skin lesions: comparison of the abcd rule of dermatoscopy and a new 7-point
  checklist based on pattern analysis.
\newblock {\em Archives of dermatology}, 134(12):1563--1570, 1998.

\bibitem{abcd}
Giuseppe Argenziano, Iris Zalaudek, and H.~Peter Soyer.
\newblock Which is the most reliable method for teaching dermoscopy for
  melanoma diagnosis to residents in dermatology?
\newblock {\em British Journal of Dermatology}, 151, 2004.

\bibitem{logic_CBM}
Pietro Barbiero, Gabriele Ciravegna, Francesco Giannini, Pietro Li{\'o}, Marco
  Gori, and Stefano Melacci.
\newblock Entropy-based logic explanations of neural networks.
\newblock In {\em Proceedings of the AAAI Conference on Artificial
  Intelligence}, volume~36, pages 6046--6054, 2022.

\bibitem{bias2}
Alceu Bissoto, Michel Fornaciali, Eduardo Valle, and Sandra Avila.
\newblock ({D}e){C}onstructing bias on skin lesion datasets.
\newblock In {\em ISIC Skin Image Anaylsis Workshop, 2019 {IEEE} Conference on
  Computer Vision and Pattern Recognition Workshops (CVPRW)}, 2019.

\bibitem{fourier}
Ronald~Newbold Bracewell and Ronald~N Bracewell.
\newblock {\em The Fourier transform and its applications}, volume 31999.
\newblock McGraw-Hill New York, 1986.

\bibitem{sklearn_api}
Lars Buitinck, Gilles Louppe, Mathieu Blondel, Fabian Pedregosa, Andreas
  Mueller, Olivier Grisel, Vlad Niculae, Peter Prettenhofer, Alexandre
  Gramfort, Jaques Grobler, Robert Layton, Jake VanderPlas, Arnaud Joly, Brian
  Holt, and Ga{\"{e}}l Varoquaux.
\newblock {API} design for machine learning software: experiences from the
  scikit-learn project.
\newblock In {\em ECML PKDD Workshop: Languages for Data Mining and Machine
  Learning}, pages 108--122, 2013.

\bibitem{isic2017}
Noel~CF Codella, David Gutman, M~Emre Celebi, Brian Helba, Michael~A Marchetti,
  Stephen~W Dusza, Aadi Kalloo, Konstantinos Liopyris, Nabin Mishra, Harald
  Kittler, et~al.
\newblock Skin lesion analysis toward melanoma detection: A challenge at the
  2017 international symposium on biomedical imaging (isbi), hosted by the
  international skin imaging collaboration (isic).
\newblock In {\em 2018 IEEE 15th international symposium on biomedical imaging
  (ISBI 2018)}, pages 168--172. IEEE, 2018.

\bibitem{ddi}
Roxana Daneshjou, Kailas Vodrahalli, Roberto~A Novoa, Melissa Jenkins, Weixin
  Liang, Veronica Rotemberg, Justin Ko, Susan~M Swetter, Elizabeth~E Bailey,
  Olivier Gevaert, et~al.
\newblock Disparities in dermatology ai performance on a diverse, curated
  clinical image set.
\newblock {\em Science advances}, 8(31):eabq6147, 2022.

\bibitem{skincon}
Roxana Daneshjou, Mert Yuksekgonul, Zhuo~Ran Cai, Roberto~A Novoa, and James
  Zou.
\newblock Skincon: A skin disease dataset densely annotated by domain experts
  for fine-grained debugging and analysis.
\newblock In {\em Thirty-sixth Conference on Neural Information Processing
  Systems Datasets and Benchmarks Track}, 2022.

\bibitem{human_attention}
Abhishek Das, Harsh Agrawal, Larry Zitnick, Devi Parikh, and Dhruv Batra.
\newblock Human attention in visual question answering: Do humans and deep
  networks look at the same regions?
\newblock {\em Computer Vision and Image Understanding}, 163:90--100, 2017.

\bibitem{vit}
Alexey Dosovitskiy, Lucas Beyer, Alexander Kolesnikov, Dirk Weissenborn,
  Xiaohua Zhai, Thomas Unterthiner, Mostafa Dehghani, Matthias Minderer, Georg
  Heigold, Sylvain Gelly, Jakob Uszkoreit, and Neil Houlsby.
\newblock An image is worth 16x16 words: Transformers for image recognition at
  scale.
\newblock In {\em International Conference on Learning Representations}, 2021.

\bibitem{deepderm}
Andre Esteva, Brett Kuprel, Roberto~A. Novoa, Justin~M. Ko, Susan~M. Swetter,
  Helen~M. Blau, and Sebastian Thrun.
\newblock Dermatologist-level classification of skin cancer with deep neural
  networks.
\newblock {\em Nature}, 542:115--118, 2017.

\bibitem{synthderm}
Asma Ghandeharioun, Been Kim, Chun-Liang Li, Brendan Jou, Brian Eoff, and
  Rosalind~W Picard.
\newblock Dissect: Disentangled simultaneous explanations via concept
  traversals.
\newblock {\em arXiv preprint arXiv:2105.15164}, 2021.

\bibitem{f17k}
Matt Groh, Caleb Harris, Luis~R. Soenksen, Felix Lau, Rachel Han, Aerin Kim,
  Arash Koochek, and Omar Badri.
\newblock Evaluating deep neural networks trained on clinical images in
  dermatology with the fitzpatrick 17k dataset.
\newblock {\em 2021 IEEE/CVF Conference on Computer Vision and Pattern
  Recognition Workshops (CVPRW)}, pages 1820--1828, 2021.

\bibitem{isic2016}
David Gutman, Noel~CF Codella, Emre Celebi, Brian Helba, Michael Marchetti,
  Nabin Mishra, and Allan Halpern.
\newblock Skin lesion analysis toward melanoma detection: A challenge at the
  international symposium on biomedical imaging (isbi) 2016, hosted by the
  international skin imaging collaboration (isic).
\newblock {\em arXiv preprint arXiv:1605.01397}, 2016.

\bibitem{resnet}
Kaiming He, X. Zhang, Shaoqing Ren, and Jian Sun.
\newblock Deep residual learning for image recognition.
\newblock {\em 2016 IEEE Conference on Computer Vision and Pattern Recognition
  (CVPR)}, pages 770--778, 2016.

\bibitem{interactive_attention}
Jay Heo, Junhyeon Park, Hyewon Jeong, Kwang~Joon Kim, Juho Lee, Eunho Yang, and
  Sung~Ju Hwang.
\newblock Cost-effective interactive attention learning with neural attention
  processes.
\newblock In {\em International Conference on Machine Learning}, pages
  4228--4238. PMLR, 2020.

\bibitem{no_exp}
Sarthak Jain and Byron~C. Wallace.
\newblock Attention is not explanation.
\newblock {\em NAACL 2019}, abs/1902.10186, 2019.

\bibitem{derm7pt}
Jeremy Kawahara, Sara Daneshvar, Giuseppe Argenziano, and Ghassan Hamarneh.
\newblock Seven-point checklist and skin lesion classification using multitask
  multimodal neural nets.
\newblock {\em IEEE Journal of Biomedical and Health Informatics},
  23(2):538--546, mar 2019.

\bibitem{cav}
Been Kim, Martin Wattenberg, Justin Gilmer, Carrie~J. Cai, James Wexler,
  Fernanda~B. Vi{\'e}gas, and Rory Sayres.
\newblock Interpretability beyond feature attribution: Quantitative testing
  with concept activation vectors (tcav).
\newblock In {\em ICML}, 2018.

\bibitem{CBM}
Pang~Wei Koh, Thao Nguyen, Yew~Siang Tang, Stephen Mussmann, Emma Pierson, Been
  Kim, and Percy Liang.
\newblock Concept bottleneck models.
\newblock In {\em International Conference on Machine Learning}, pages
  5338--5348. PMLR, 2020.

\bibitem{spay}
Sebastian Lapuschkin, Stephan W{\"a}ldchen, Alexander Binder, Gr{\'e}goire
  Montavon, Wojciech Samek, and Klaus-Robert M{\"u}ller.
\newblock Unmasking clever hans predictors and assessing what machines really
  learn.
\newblock {\em Nature communications}, 10(1):1--8, 2019.

\bibitem{skin_cav}
Adriano Lucieri, Muhammad~Naseer Bajwa, Stephan~Alexander Braun, Muhammad~Imran
  Malik, Andreas Dengel, and Sheraz Ahmed.
\newblock On interpretability of deep learning based skin lesion classifiers
  using concept activation vectors.
\newblock In {\em 2020 international joint conference on neural networks
  (IJCNN)}, pages 1--10. IEEE, 2020.

\bibitem{bias3}
Agnieszka Miko{\l}ajczyk, Sylwia Majchrowska, and Sandra Carrasco~Limeros.
\newblock The (de) biasing effect of gan-based augmentation methods on skin
  lesion images.
\newblock In {\em International Conference on Medical Image Computing and
  Computer-Assisted Intervention}, pages 437--447. Springer, 2022.

\bibitem{lime}
Kewen Peng and Tim Menzies.
\newblock Documenting evidence of a reuse of ‘“why should i trust you?”:
  explaining the predictions of any classifier’.
\newblock In {\em Proceedings of the 29th ACM Joint Meeting on European
  Software Engineering Conference and Symposium on the Foundations of Software
  Engineering}, pages 1600--1600, 2021.

\bibitem{bias4}
Samuel~William Pewton and Moi~Hoon Yap.
\newblock Dark corner on skin lesion image dataset: Does it matter?
\newblock In {\em Proceedings of the IEEE/CVF Conference on Computer Vision and
  Pattern Recognition (CVPR) Workshops}, pages 4831--4839, June 2022.

\bibitem{cd}
Laura Rieger, Chandan Singh, William Murdoch, and Bin Yu.
\newblock Interpretations are useful: penalizing explanations to align neural
  networks with prior knowledge.
\newblock In {\em International conference on machine learning}, pages
  8116--8126. PMLR, 2020.

\bibitem{inter_align}
Laura Rieger, Chandan Singh, W.~James Murdoch, and Bin Yu.
\newblock Interpretations are useful: Penalizing explanations to align neural
  networks with prior knowledge.
\newblock In {\em Proceedings of the 37th International Conference on Machine
  Learning, {ICML} 2020, 13-18 July 2020, Virtual Event}, volume 119 of {\em
  Proceedings of Machine Learning Research}, pages 8116--8126. {PMLR}, 2020.

\bibitem{CTransformer}
Mattia Rigotti, Christoph Miksovic, Ioana Giurgiu, Thomas Gschwind, and Paolo
  Scotton.
\newblock Attention-based interpretability with concept transformers.
\newblock In {\em The Tenth International Conference on Learning
  Representations, {ICLR} 2022, Virtual Event, April 25-29, 2022}.
  OpenReview.net, 2022.

\bibitem{rrr}
Andrew~Slavin Ross, Michael~C. Hughes, and Finale Doshi{-}Velez.
\newblock Right for the right reasons: Training differentiable models by
  constraining their explanations.
\newblock In Carles Sierra, editor, {\em Proceedings of the Twenty-Sixth
  International Joint Conference on Artificial Intelligence, {IJCAI} 2017,
  Melbourne, Australia, August 19-25, 2017}, pages 2662--2670. ijcai.org, 2017.

\bibitem{rrr_nmi}
Patrick Schramowski, Wolfgang Stammer, Stefano Teso, Anna Brugger, Franziska
  Herbert, Xiaoting Shao, Hans{-}Georg Luigs, Anne{-}Katrin Mahlein, and
  Kristian Kersting.
\newblock Making deep neural networks right for the right scientific reasons by
  interacting with their explanations.
\newblock {\em Nat. Mach. Intell.}, 2(8):476--486, 2020.

\bibitem{xil_nature}
Patrick Schramowski, Wolfgang Stammer, Stefano Teso, Anna Brugger, Franziska
  Herbert, Xiaoting Shao, Hans{-}Georg Luigs, Anne{-}Katrin Mahlein, and
  Kristian Kersting.
\newblock Making deep neural networks right for the right scientific reasons by
  interacting with their explanations.
\newblock {\em Nat. Mach. Intell.}, 2(8):476--486, 2020.

\bibitem{gradcam}
Ramprasaath~R. Selvaraju, Michael Cogswell, Abhishek Das, Ramakrishna Vedantam,
  Devi Parikh, and Dhruv Batra.
\newblock Grad-cam: Visual explanations from deep networks via gradient-based
  localization.
\newblock {\em Int. J. Comput. Vis.}, 128(2):336--359, 2020.

\bibitem{vgg}
Karen Simonyan and Andrew Zisserman.
\newblock Very deep convolutional networks for large-scale image recognition.
\newblock {\em CoRR}, abs/1409.1556, 2015.

\bibitem{Nesy}
Wolfgang Stammer, Patrick Schramowski, and Kristian Kersting.
\newblock Right for the right concept: Revising neuro-symbolic concepts by
  interacting with their explanations.
\newblock In {\em {IEEE} Conference on Computer Vision and Pattern Recognition,
  {CVPR} 2021, virtual, June 19-25, 2021}, pages 3619--3629. Computer Vision
  Foundation / {IEEE}, 2021.

\bibitem{ig}
Mukund Sundararajan, Ankur Taly, and Qiqi Yan.
\newblock Axiomatic attribution for deep networks.
\newblock In {\em International conference on machine learning}, pages
  3319--3328. PMLR, 2017.

\bibitem{inception}
Christian Szegedy, Vincent Vanhoucke, Sergey Ioffe, Jonathon Shlens, and
  Zbigniew Wojna.
\newblock Rethinking the inception architecture for computer vision.
\newblock {\em 2016 IEEE Conference on Computer Vision and Pattern Recognition
  (CVPR)}, pages 2818--2826, 2016.

\bibitem{isic2019}
Philipp Tschandl, Cliff Rosendahl, and Harald Kittler.
\newblock The ham10000 dataset, a large collection of multi-source
  dermatoscopic images of common pigmented skin lesions.
\newblock {\em Scientific Data}, 5, 2018.

\bibitem{tsne}
Laurens van~der Maaten and Geoffrey Hinton.
\newblock Visualizing data using t-sne.
\newblock {\em Journal of Machine Learning Research}, 9(86):2579--2605, 2008.

\bibitem{isic2020}
Rotemberg Veronica, Kurtansky Nicholas, Betz-Stablein Brigid, Caffery Liam,
  Chousakos Emmanouil, Codella Noel, Combalia Marc, Stephen Dusza, Guitera
  Pascale, David Gutman, et~al.
\newblock A patient-centric dataset of images and metadata for identifying
  melanomas using clinical context.
\newblock {\em Scientific Data}, 8(1), 2021.

\bibitem{spectral}
Ulrike Von~Luxburg.
\newblock A tutorial on spectral clustering.
\newblock {\em Statistics and computing}, 17(4):395--416, 2007.

\bibitem{bias1}
Fei Wang, Lawrence Casalino, and Dhruv Khullar.
\newblock Deep learning in medicine—promise, progress, and challenges.
\newblock {\em JAMA Internal Medicine}, 179, 12 2018.

\bibitem{pcbm}
Mert Yuksekgonul, Maggie Wang, and James Zou.
\newblock Post-hoc concept bottleneck models.
\newblock In {\em ICLR 2022 Workshop on PAIR{\textasciicircum}2Struct: Privacy,
  Accountability, Interpretability, Robustness, Reasoning on Structured Data},
  2022.

\bibitem{cam}
Bolei Zhou, Aditya Khosla, Agata Lapedriza, Aude Oliva, and Antonio Torralba.
\newblock Learning deep features for discriminative localization.
\newblock In {\em Proceedings of the IEEE conference on computer vision and
  pattern recognition}, pages 2921--2929, 2016.

\end{thebibliography}
}

\begin{abstract}
In this supplementary material, we provide more details
about datasets, additional training details, network architectures, t-SNE visualisation, concept accuracy, and explanation visualisation.
\end{abstract}
\section{More Details about Datasets}
\subsection{General Datasets}
We choose \textit{SynthDerm}, \textit{ISIC2016}, \textit{ISIC2017}, and \textit{ISIC2019\_2020} as evaluating datasets in the section ``Confounding Concept Discovery" of the experimental part. We chose SynthDerm as it is a well-controlled dataset and chose other datasets due to their popularity in dermatology. Also, we choose \textit{Fitzpatrick17k} and \textit{DDI} as training and testing dataset in the section `` Debiasing the Negative Impact of Skin Tone" as they contain rich Fitzpatrick skin type labels.\\
\noindent\textbf{SynthDerm:} \textit{SynthDerm} \cite{synthderm} is a balanced synthetic dataset inspired by real-world ABCD rule criteria \cite{abcd} of melanoma skin lesions. It includes images with different factors, including whether asymmetric, different borders, colors, diameter, or evolving in size, shape, and color over time. For skin tone, it simulates six Fitzpatrick skin scales. It includes 2600 64x64 images. Moreover, in this dataset, there are surgical markings in melanoma images but not in benign images. Thus, the ``surgical markings" is the confounding factors in the dataset.
\\
\noindent\textbf{ISIC2016:} We use the data from the task 3 of \textit{ISIC2016} \cite{isic2016} challenge, it contains 900 dermoscopic images. \\
\noindent\textbf{ISIC2017:} We use the data from the part 3 of \textit{ISIC2017} \cite{isic2017} challenge, it contains 2000 dermoscopic images.\\
\noindent\textbf{ISIC2019\_2020:} \textit{ISIC2019\_2020} \cite{isic2019,isic2020} is the \textit{ISIC2020} dataset with all melanoma images from \textit{ISIC2019}, which includes 37648 dermoscopic images.\\
\noindent\textbf{Fitzpatrick17k:} \textit{Fitzpatrick17k} \cite{f17k} contains 16577 clinical images labeled by 114 skin conditions and 6 Fitzpatrick skin types. \\
\noindent\textbf{DDI:} \textit{DDI} \cite{ddi} is similar to \textit{Fitzpatrick17k} but with higher quality. It contains 208 images of FST (\uppercase\expandafter{\romannumeral1}-\uppercase\expandafter{\romannumeral2}), 241 images of FST (\uppercase\expandafter{\romannumeral3-\uppercase\expandafter{\romannumeral4}}), and 207 images of FST (\uppercase\expandafter{\romannumeral1-\uppercase\expandafter{\romannumeral6}}). which corresponds to light skin, middle skin, and dark skin tone, respectively.

\subsection{Probe Datasets:}
For constructing the concept bank, we use \textit{Derm7pt} as the probe dataset for dermoscopic image dataset such as \textit{ConfDerm} and use \textit{SKINCON} as the probe dataset for clinical image dataset such as \textit{Fitzpatrick17k}.

\noindent\textbf{SKINCON:} \textit{SKINCON} \cite{skincon} is a  skin disease dataset densely annotated by domain experts for fine-grained model debugging and analysis. It includes 3230 images with 48 clinical concepts, 22 of which have over 50 images.\\
\noindent\textbf{Derm7pt:} \textit{Derm7pt} \cite{derm7pt} is a dermoscopic image dataset contains 1011 dermoscopic images with 7 clinical concepts (\ie{pigmentation network, blue whitish veil, vascular structures, pigmentation, streaks, dots and globules, and regression structures.})\cite{7pt} for melanoma skin lesions in dermatology.

\subsection{ConfDerm:}
\begin{figure*}[t]
\setlength{\abovecaptionskip}{0.cm}
   \begin{center}
     \renewcommand{\arraystretch}{0.2}
   \begin{tabular}{ c@{ } c@{ } c@{ } c@{ }  c@{ }  c@{ } c@{ }  }

   {\includegraphics[width=0.18\linewidth, height=0.13\linewidth]{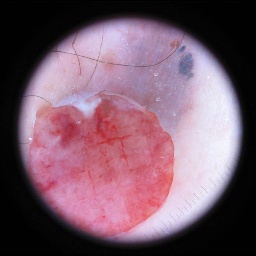}}&
   {\includegraphics[width=0.18\linewidth, height=0.13\linewidth]{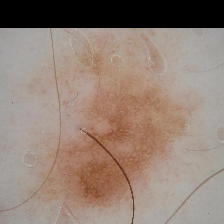}}&
   {\includegraphics[width=0.18\linewidth, height=0.13\linewidth]{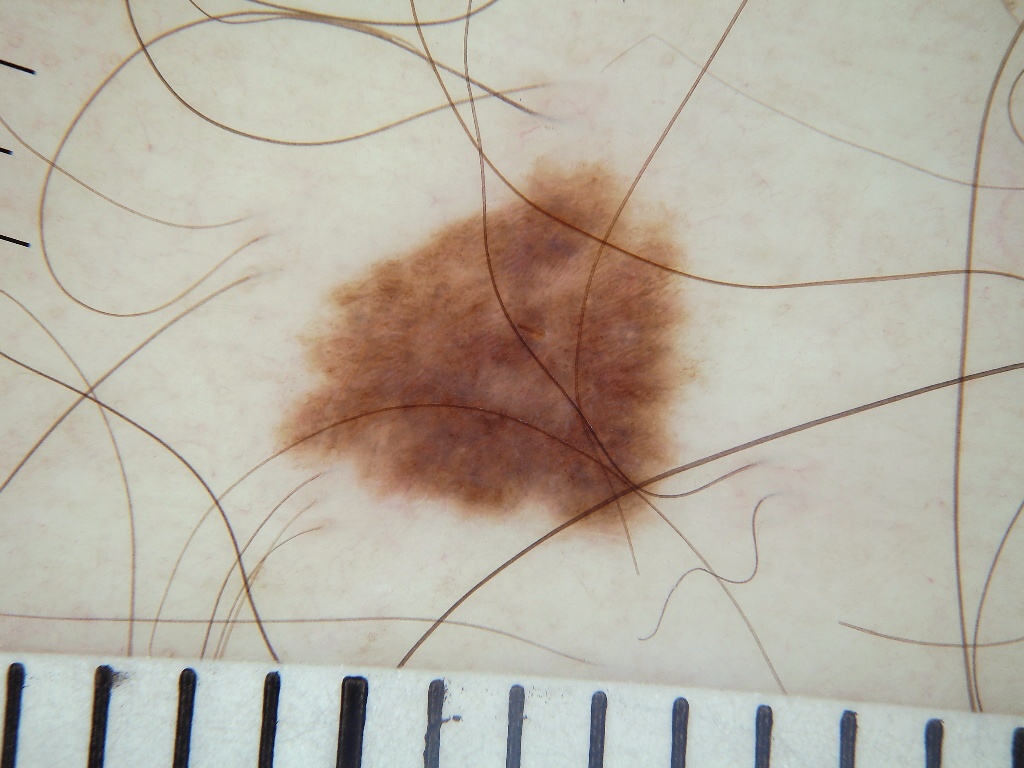}}&
   {\includegraphics[width=0.18\linewidth, height=0.13\linewidth]{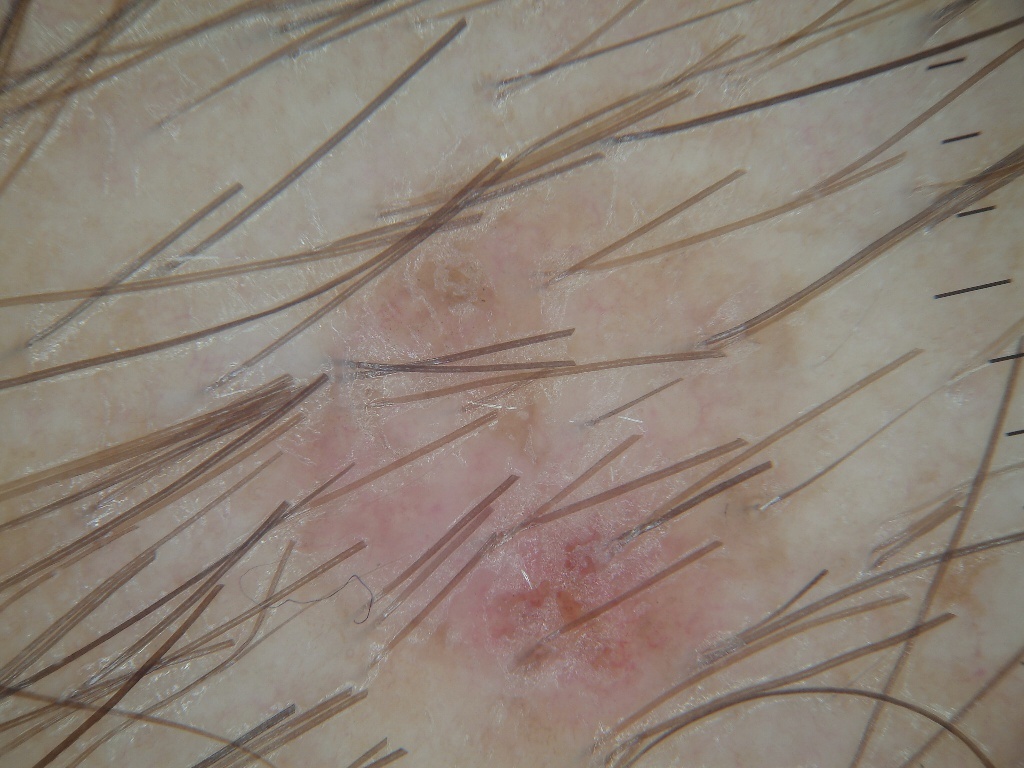}}&
   {\includegraphics[width=0.18\linewidth, height=0.13\linewidth]{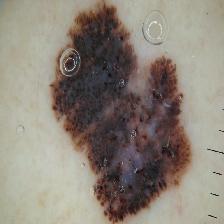}}
   
   \\
      {\includegraphics[width=0.18\linewidth, height=0.13\linewidth]{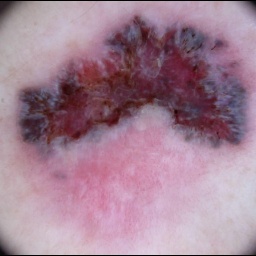}}&
   {\includegraphics[width=0.18\linewidth, height=0.13\linewidth]{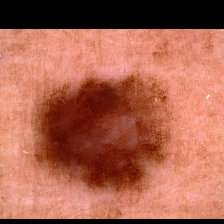}}&
   {\includegraphics[width=0.18\linewidth, height=0.13\linewidth]{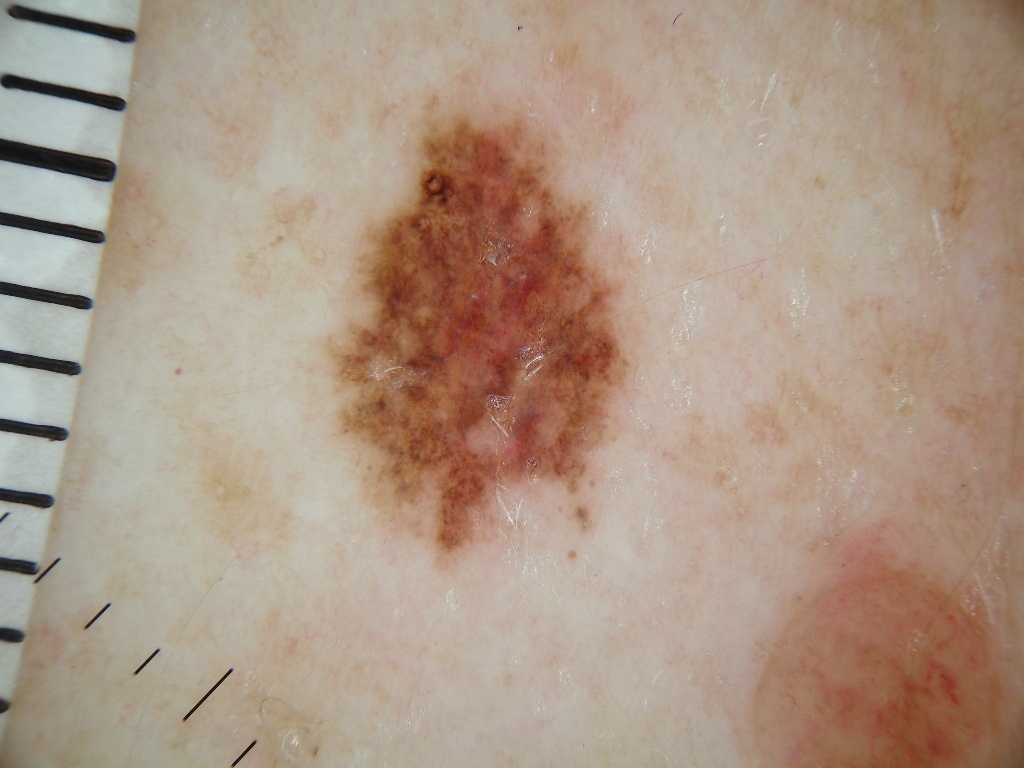}}&
   {\includegraphics[width=0.18\linewidth, height=0.13\linewidth]{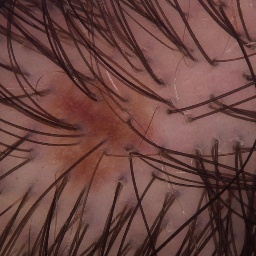}}&
   {\includegraphics[width=0.18\linewidth, height=0.13\linewidth]{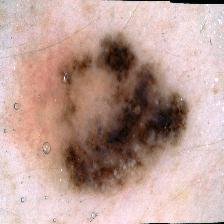}}
   
   \\

      {\includegraphics[width=0.18\linewidth, height=0.13\linewidth]{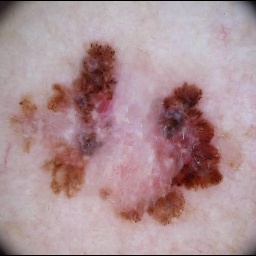}}&
   {\includegraphics[width=0.18\linewidth, height=0.13\linewidth]{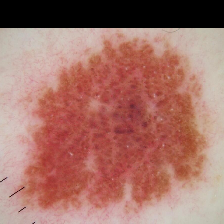}}&
   {\includegraphics[width=0.18\linewidth, height=0.13\linewidth]{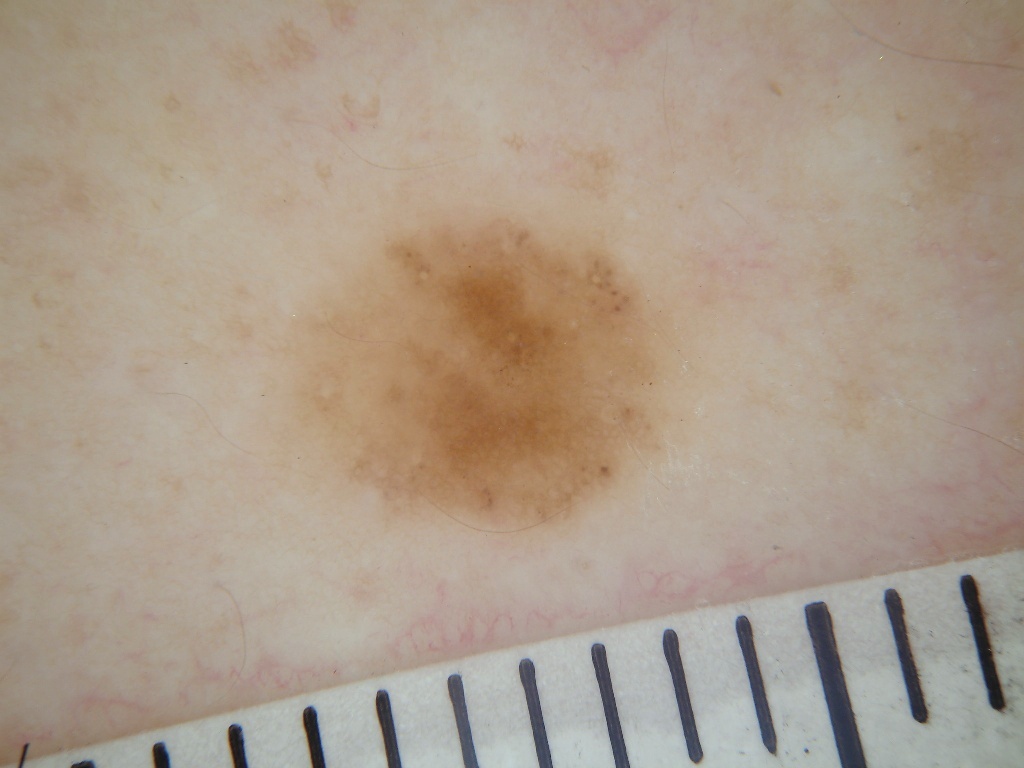}}&
   {\includegraphics[width=0.18\linewidth, height=0.13\linewidth]{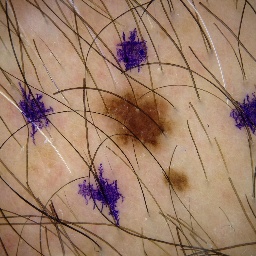}}&
   {\includegraphics[width=0.18\linewidth, height=0.13\linewidth]{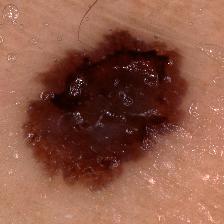}}
   
   \\

    {\includegraphics[width=0.18\linewidth, height=0.13\linewidth]{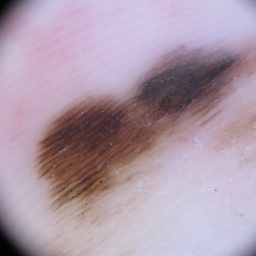}}&
   {\includegraphics[width=0.18\linewidth, height=0.13\linewidth]{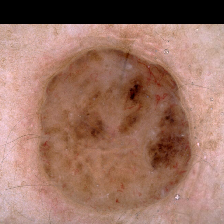}}&
   {\includegraphics[width=0.18\linewidth, height=0.13\linewidth]{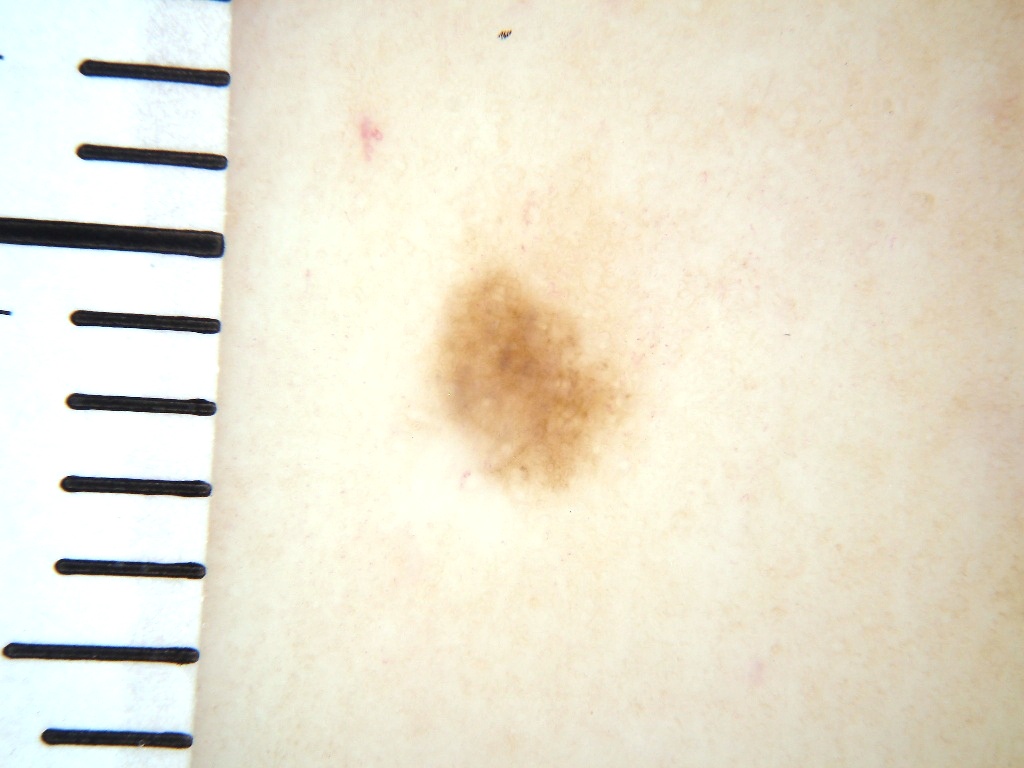}}&
   {\includegraphics[width=0.18\linewidth, height=0.13\linewidth]{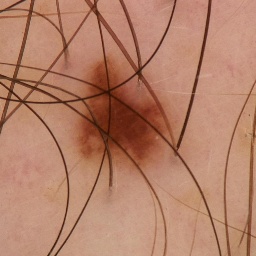}}&
   {\includegraphics[width=0.18\linewidth, height=0.13\linewidth]{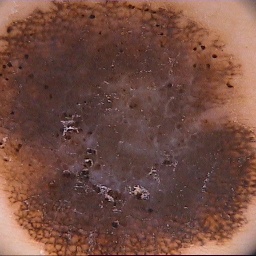}}
   
   \\

      {\includegraphics[width=0.18\linewidth, height=0.13\linewidth]{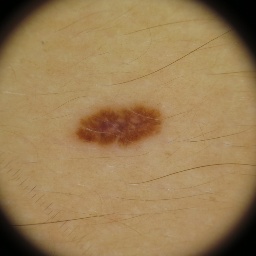}}&
   {\includegraphics[width=0.18\linewidth, height=0.13\linewidth]{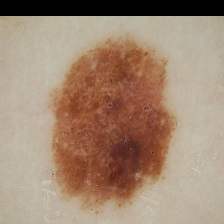}}&
   {\includegraphics[width=0.18\linewidth, height=0.13\linewidth]{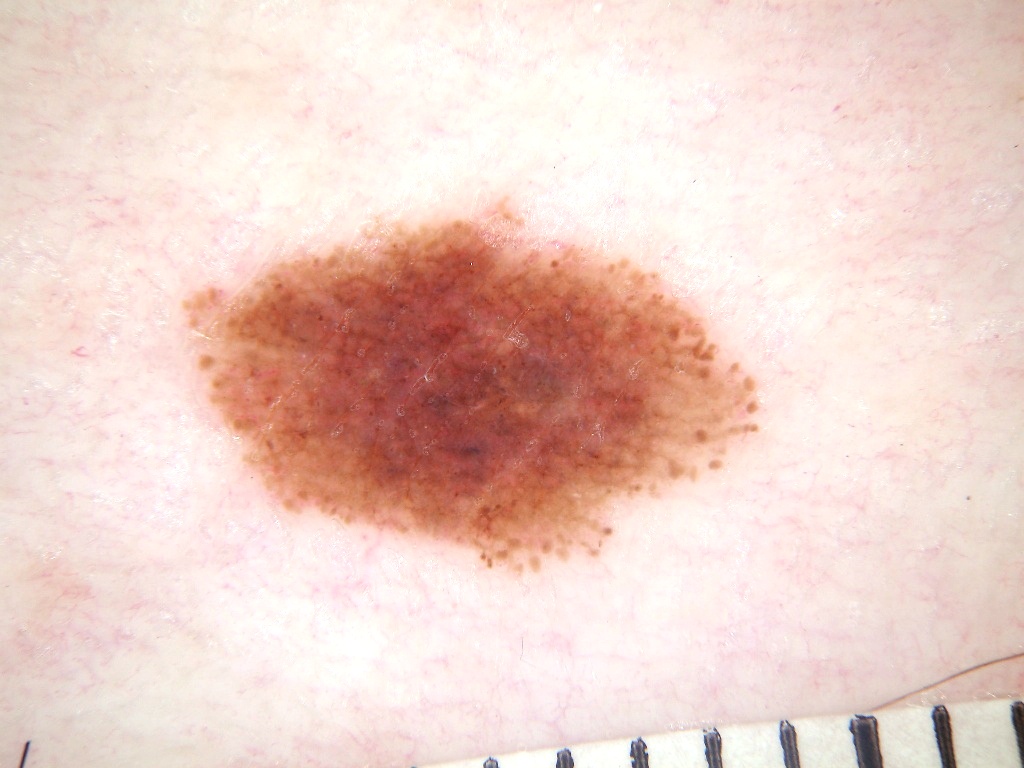}}&
    {\includegraphics[width=0.18\linewidth, height=0.13\linewidth]{supp/aps/ISIC_0000310_mel.jpg}}&
   {\includegraphics[width=0.18\linewidth, height=0.13\linewidth]{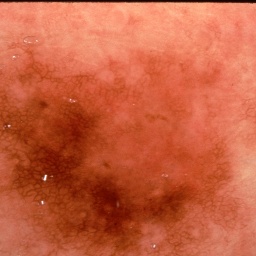}}
      \\
 
      {\includegraphics[width=0.18\linewidth, height=0.13\linewidth]{conf_imgs/dc_img.png}}&
   {\includegraphics[width=0.18\linewidth, height=0.13\linewidth]{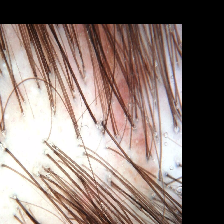}}&
   {\includegraphics[width=0.18\linewidth, height=0.13\linewidth]{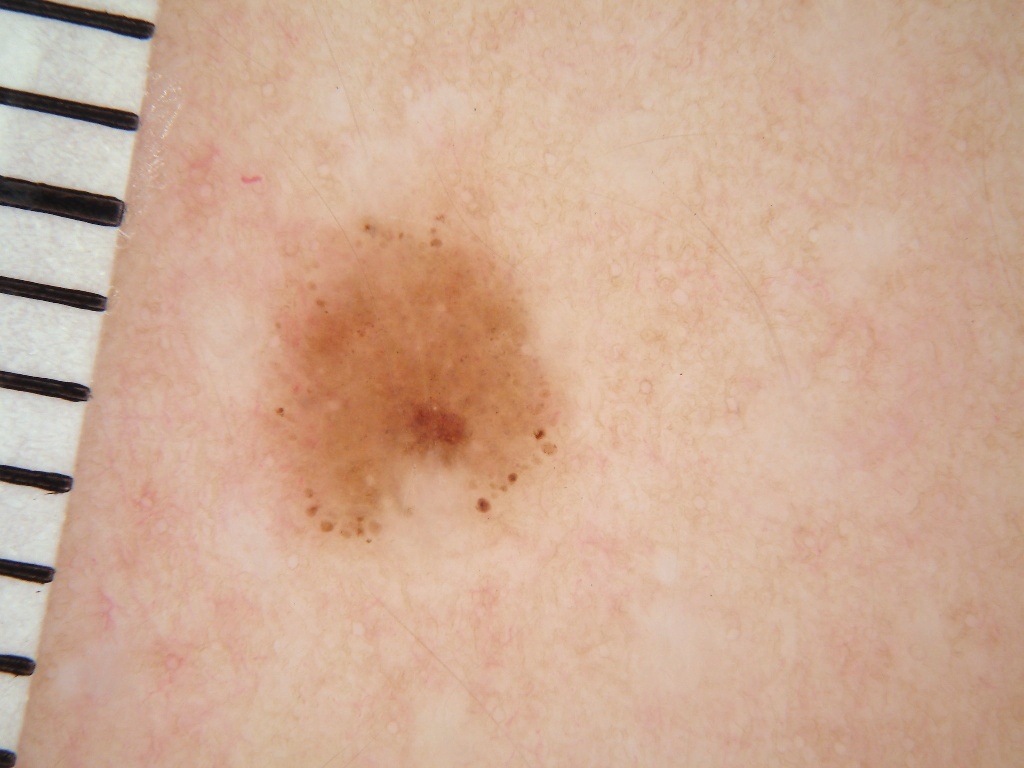}}&
   {\includegraphics[width=0.18\linewidth, height=0.13\linewidth]{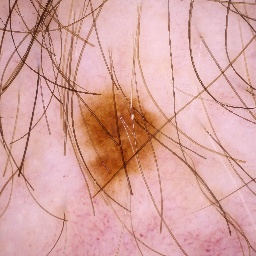}}&
   {\includegraphics[width=0.18\linewidth, height=0.13\linewidth]{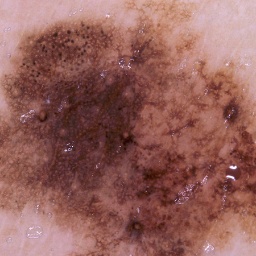}}
      \\

      {\includegraphics[width=0.18\linewidth, height=0.13\linewidth]{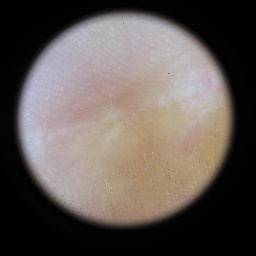}}&
   {\includegraphics[width=0.18\linewidth, height=0.13\linewidth]{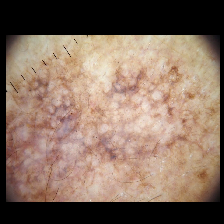}}&
   {\includegraphics[width=0.18\linewidth, height=0.13\linewidth]{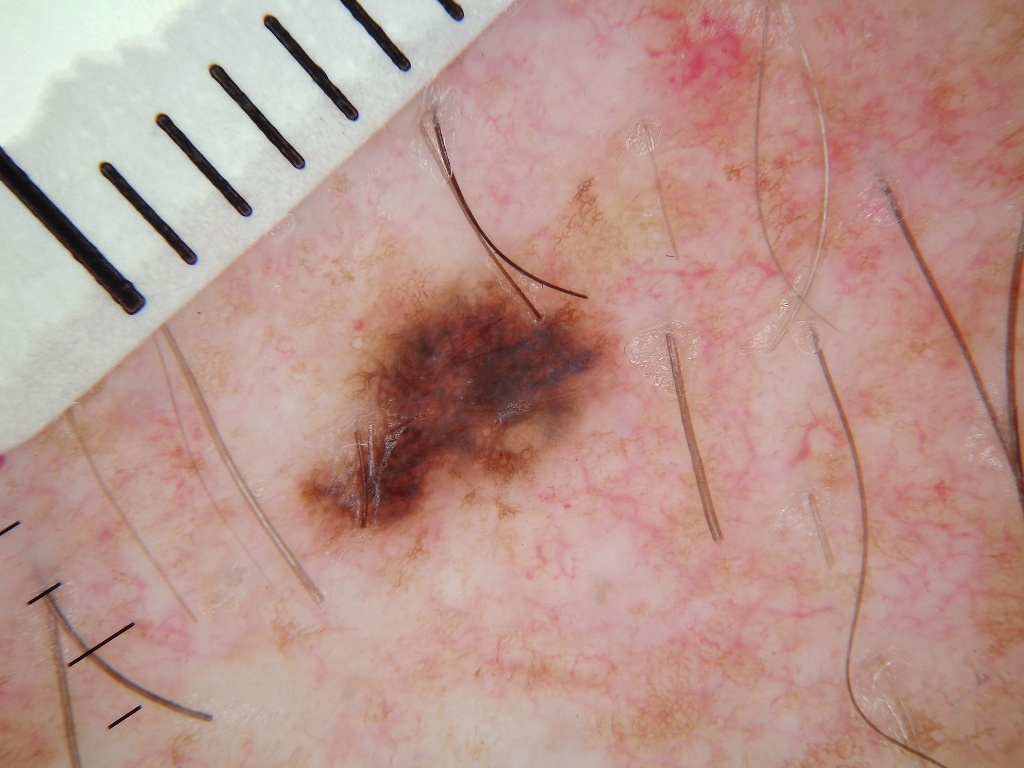}}&
   {\includegraphics[width=0.18\linewidth, height=0.13\linewidth]{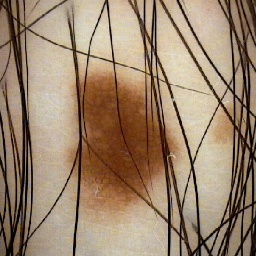}}&
   {\includegraphics[width=0.18\linewidth, height=0.13\linewidth]{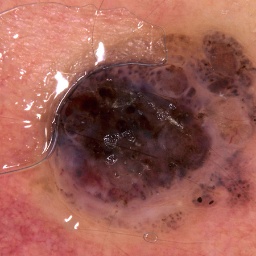}}
   
   \\
   \footnotesize{melanoma (dark corners)} & \footnotesize{benign (dark borders)} & \footnotesize{benign (rulers)} & \footnotesize{benign (hairs)} & \footnotesize{melanoma (air pockets)}  \\
   \end{tabular}
      \caption{Visualization of the confounded class of five datasets in our \textit{ConfDerm}, each one has one confounding factor, including dark corners, dark borders, hairs, and air pockets.}
      \vspace{-2mm}
      \label{fig9}
   \end{center}
%   \setlength{\abovecaptionskip}{0.cm}
% \setlength{\belowcaptionskip}{-0.cm}
%   \vspace{2mm}
\end{figure*}

We provide additional data visualization, showing the characteristics of images in the confounded class of five datasets in our ConfDerm dataset, as illustrated in Fig. \ref{fig9}.

\section{Additional Training Details and Network Architecture}

\begin{figure}[!t]
 \vspace{-2mm}
   \begin{center}
   {\includegraphics[width=0.95\linewidth,height=0.5\linewidth]{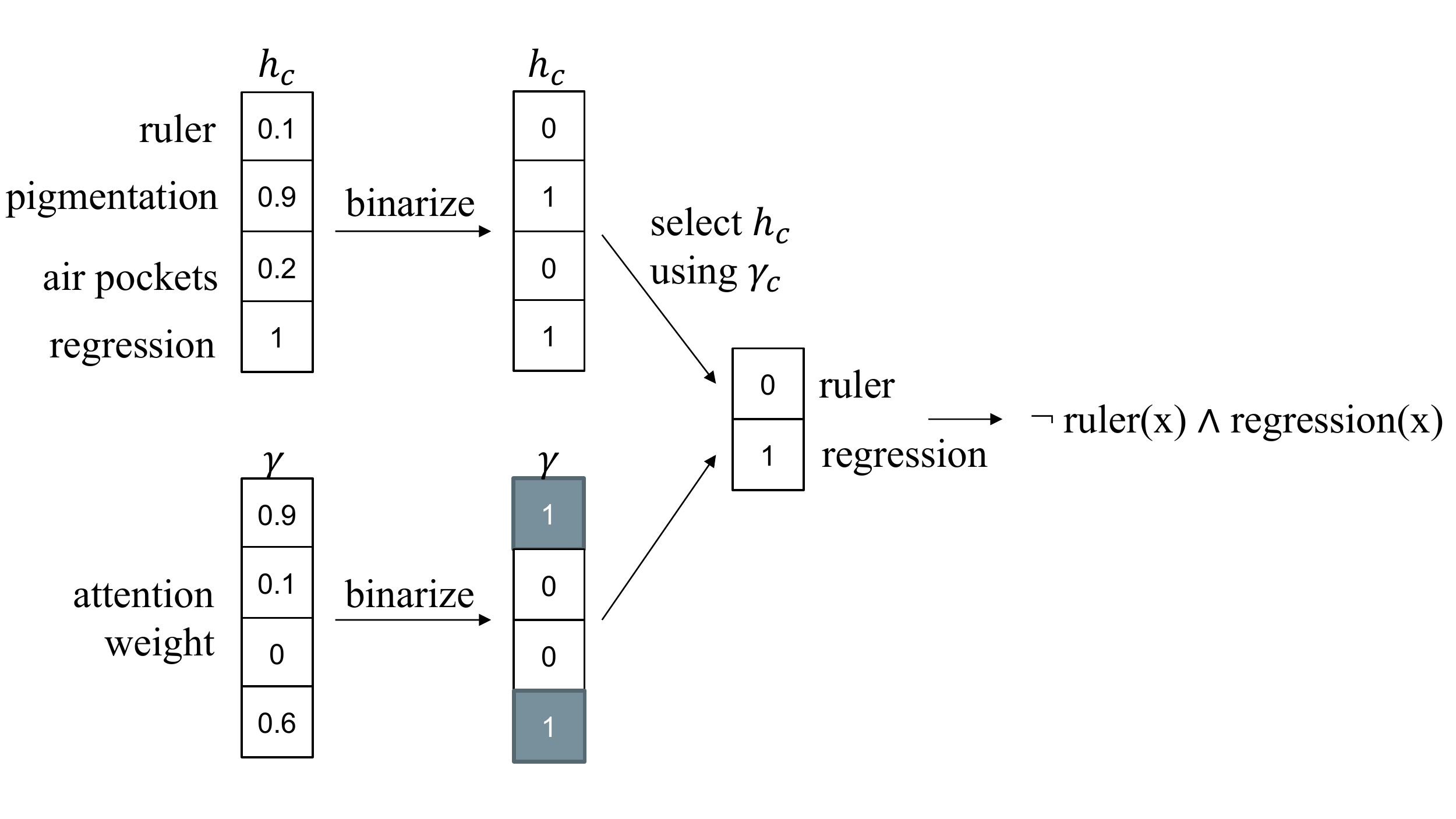}}
   \end{center}
   \vspace{-5mm}
\caption{\small Illustration of the logical explanation generation. }
\vspace{-2mm}
   \label{supp:logic}
\end{figure}

\noindent\textbf{Detail of the logic layer:} We choose the recently proposed entropy-based logical layer \cite{logic_CBM}. It consists of four steps:  
  (1) For each concept, calculate the concept importance score $\gamma_{j}$ via calculating the $l2$ norm of all neurons in subsequent layers connected to the concept. 
  (2) Perform softmax and rescaling on the $\gamma$. 
  (3) Get the importance-aware concept score $\hat{h^{c}}$ via weighting the $\gamma$ on all concept scores $h^{c}$. 
  (4) Finally, feed the $\hat{h^{c}}$ into subsequent layers. The first-order logic generation of the model is described in the example of Fig.~\ref{supp:logic}. It binaries the concept scores $h^{c}$ and the attention weights $\gamma$, then select one concept if its weight $\gamma_{j}$ is 1. 

This method is based on attention operation, but \cite{human_attention,no_exp} shows that attention is often not the explanation, which causes interaction on it is not effective in changing the model's behavior. In Fig. \ref{supp:logic}, it shows that global explanations of the model using attention and our method, after the interaction, the left of Fig. \ref{supp:logic} shows that the model using attention still focuses on the "ruler" concept, and the right of Fig. \ref{supp:logic} shows that the model using our explanation does not give a high weight for the ruler and can focus on meaningful clinical concepts.\\

\noindent\textbf{Training Details for "Rewriting Model's Decision in ConfDerm" :} For concept bank construction, we train a linear SVM using the sklearn library \cite{sklearn_api} with regularization $\beta=0.14$ for each concept. We totally train 17 concept vectors, where 12 concepts are from the Derm7pt dataset and 5 concepts from our GCCD algorithm. All 17 concepts we obtained are \textit{``regular\textunderscore pigment\textunderscore network", ``irregular\textunderscore pigment\textunderscore network", ``blue\textunderscore whitish\textunderscore veil", ``regular\textunderscore vascular\textunderscore structures", ``irregular\textunderscore vascular\textunderscore structures", ``typicalpigmentation", ``atypical\textunderscore pigmentation", ``regular\textunderscore streaks", ``irregular\textunderscore streaks", ``regular\textunderscore dots\textunderscore and\textunderscore globules", ``irregular\textunderscore dots\textunderscore and\textunderscore globules", ``regression\textunderscore structures", ``dark corner", ``dark border", ``air pockets", ``ruler", ``hair"}.                             

For model training, we train our framework using PyTorch with a maximum of 20 epochs on each subdataset on \textit{ConfDerm} dataset.
Each image is rescaled to $256\times 256$. The black-box model is initialized with ResNet50 trained on ImageNet, and we set the logic layer using two linear layers. We use Adam optimiser and set the learning rate with 0.001, and we set the balanced weights $\lambda_{1}$ and $\lambda_{2}$ of our loss with 0.05 and 2000.\\
\noindent\textbf{Training Details for "Debiasing the Negative Impact of Skin Tone" :}
For concept bank construction, similarly,  we train a linear SVM using the sklearn library \cite{sklearn_api} with regularization $\beta=0.1$ for each concept. We choose 22 concepts that have at least 50 images and one additional confounding concept, "dark skin" from the \textit{SKINCON} dataset. To the end, all 23 concepts we collected are \textit{"Papule", "Plaque", "Pustule", "Bulla", "Patch", "Nodule", "Ulcer", "Crust", "Erosion", "Atrophy", "Exudate", "Telangiectasia' "Scale", "Scar", "Friable", "Dome-shaped", "Brown(Hyperpigmentation)", "White(Hypopigmentation)", "Purple", "Yellow", "Black", "Erythema", "dark skin"}.

For model training, we split the \textit{Fitzpatrick17k} dataset into training and validation set with a ratio of 8:2 and use \textit{DDI} dataset as the testing set. We train our framework using PyTorch with a maximum of 30 epochs and use Adam optimiser, and set the learning rate with 3e-4, and we set the balanced weights $\lambda_{1}$ and $\lambda_{2}$ of our loss with 0.1 and 4000. Each image is rescaled to $256\times 256$. The black-box model is initialized with InceptionV3 \cite{inception} trained on the dataset \cite{deepderm}. as similar to \cite{skincon}, and we set the logic layer using three linear layers.\\

\begin{figure}[!t]
  \begin{center}
  {\includegraphics[width=0.98\linewidth]{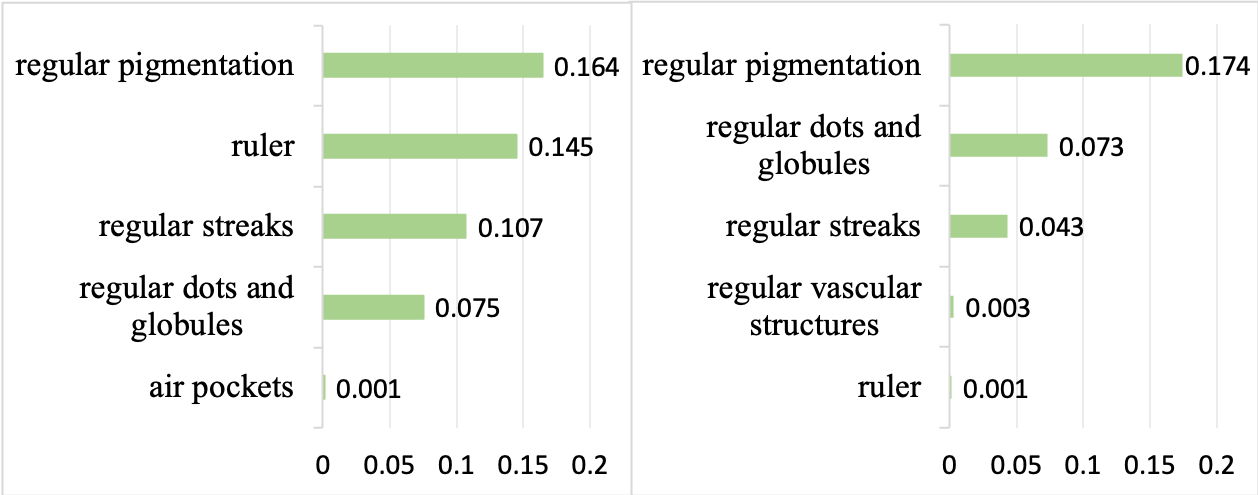}}
%   {\includegraphics[width=0.48\linewidth]{image/f5b.png}}
  \end{center}
  \vspace{-4mm}
\caption{\small Global explanation (concept activation) of reducing the confounding factor "rulers" on \textit{ConfDerm} dataset. From left to right, Interaction on attention, Interaction on our explanation. Results show that interaction with our explanation successfully alleviate the negative impact of the confounding factor.}
  \label{fig11}
  \vspace{-2mm}
\end{figure}

\section{Additional Experiments}
\subsection{More Visualisation about Confounding Concept Discovery}
\begin{figure*}[!t]
   \begin{center}
  {\includegraphics[width=0.96\linewidth]{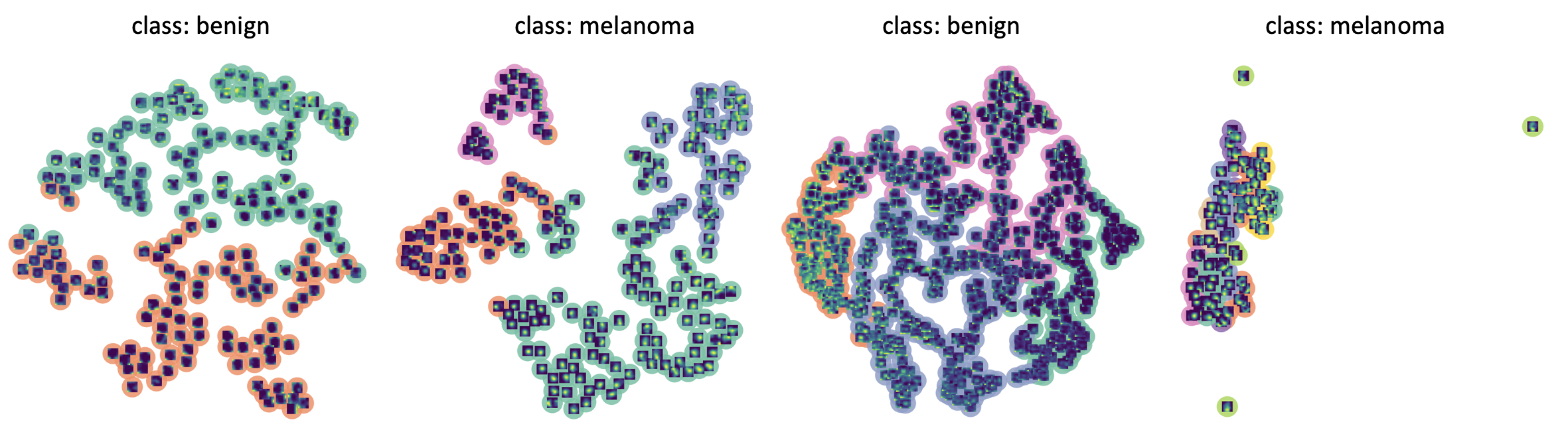}}
   \end{center}
  \vspace{-4mm}
\caption{\small Global analysis of the models' behavior within datasets using GCCD. The two left graphs are the tSNE of spectral clustering using GradCAMs of a ResNet50 within \textit{ISIC2016}. The two right are the tSNE of spectral clustering using GradCAMs of a ResNet50 within the \textit{ISIC2017} dataset.}
   \label{supp:fig4}
\end{figure*}

\noindent\textbf{GCDD on ISIC2016 and ISIC2017:} We also visualize the t-SNE of our GCDD algorithm within \textit{ISIC2016} and \textit{ISIC2017}, as shown in Fig. \ref{supp:fig4}. 

\begin{figure*}[!t]
   \begin{center}
  {\includegraphics[width=0.98\linewidth]{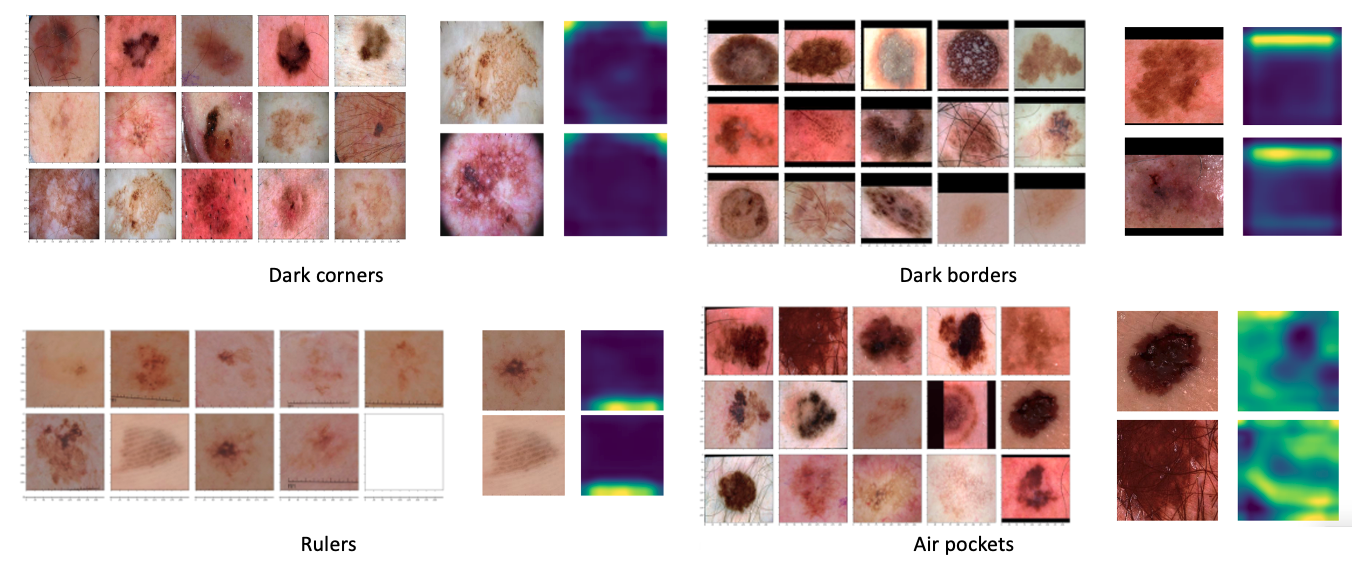}}
   \end{center}
  \vspace{-4mm}
\caption{\small Visulisation the representative clusters of GCCD on \textit{ISIC2019\textunderscore 2020.}}
   \label{fig:sample1920}
\end{figure*}

\noindent\textbf{Samples of Representative Clusters within ISIC2019\textunderscore 2020:} The representative clusters of GCDD on \textit{ISIC2019\textunderscore 2020 are illustrated in Fig. \ref{fig:sample1920} }.

\subsection{Concept Learning}
We report the testing accuracy of each concept in \textit{Derm7pt} and \textit{SKINCON} dataset, as shown in Table \ref{tab5} and Table \ref{tab6}.

\begin{figure}[!t]
  \begin{center}
  {\includegraphics[width=0.98\linewidth]{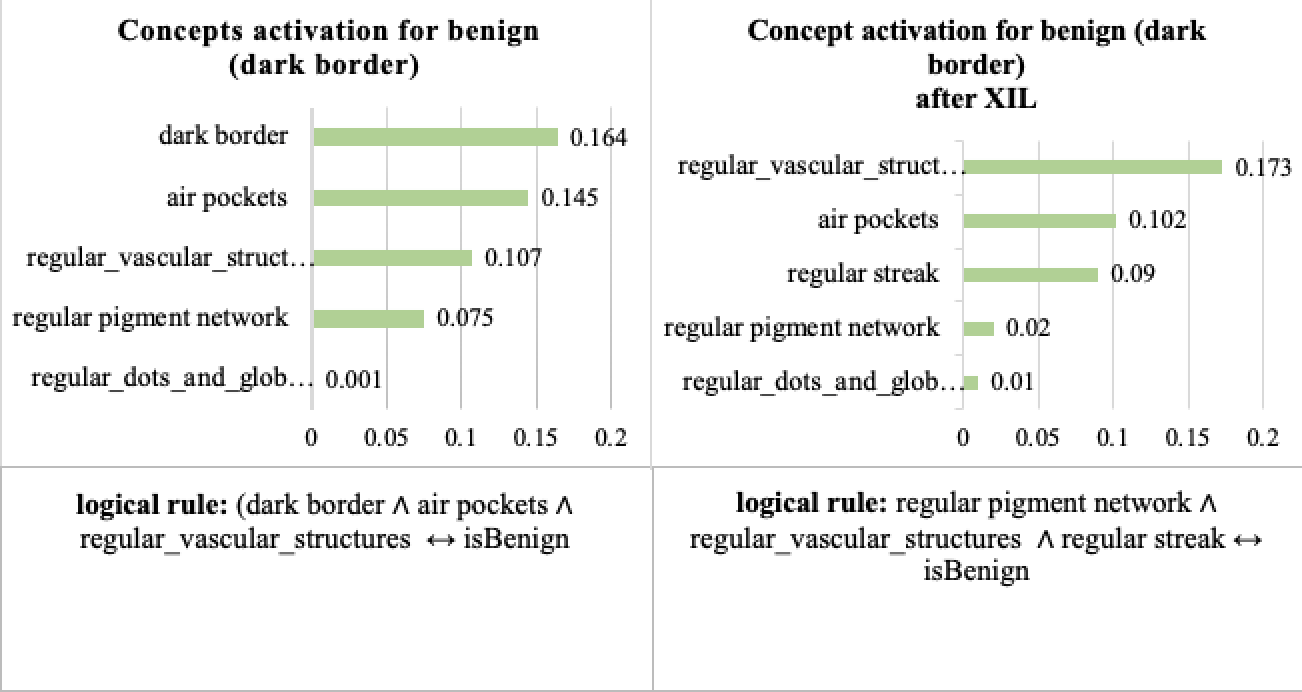}}
%   {\includegraphics[width=0.48\linewidth]{image/f5b.png}}
  \end{center}
  \vspace{-4mm}
\caption{\small The global explanation of the model's behavior on the benign (dark borders) dataset of ConfDerm. In the left figure,
either the concept activation or logical rule shows that the model is confounded by the concept of the "dark border" when predicting
benign. In the right figure, after the interaction, the model does not predict benign based on the dark corners, and it predicts
benign based on meaningful clinical concepts.}
  \label{db_xil}
  \vspace{-2mm}
\end{figure}

\begin{figure}[!t]
  \begin{center}
  {\includegraphics[width=0.98\linewidth]{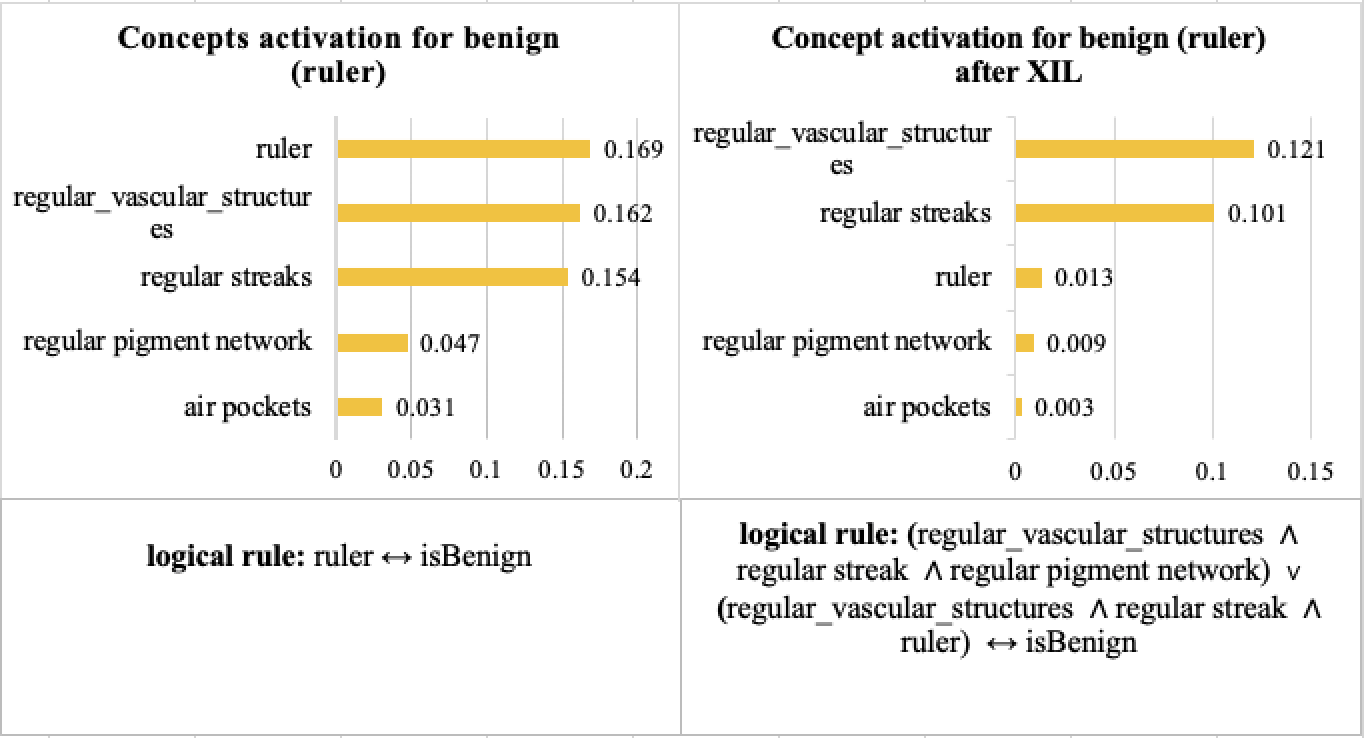}}
%   {\includegraphics[width=0.48\linewidth]{image/f5b.png}}
  \end{center}
  \vspace{-4mm}
\caption{\small The global explanation of the model's behavior on the benign (rulers) dataset of \textit{ConfDerm}. In the left figure,
either the concept activation or logical rule shows that the model is confounded by the concept of the "ruler" when predicting
benign. In the right figure, after the interaction, the model relies less on "ruler" and can predict benign based on meaningful clinical concepts.}
  \label{ruler_xil}
  \vspace{-2mm}
\end{figure}

    \begin{table}[t!]

  \centering
  \scriptsize
  \renewcommand{\arraystretch}{1.5}
  \renewcommand{\tabcolsep}{6mm}
  \caption{Concept accuracy on testing set of Derm7pt.}
    \vspace{-2mm}
  \begin{tabular}{c|c}
  \hline
  concept name & Acc (\%) \\
  \hline
regular\textunderscore pigment\textunderscore network &77.5\\  irregular\textunderscore pigment\textunderscore network &72.5\\
blue\textunderscore whitish\textunderscore veil  &70\\
regular\textunderscore vascular\textunderscore structures&70\\ irregular\textunderscore vascular\textunderscore structures  &63.33\\ typicalpigmentation &77.5\\
atypical\textunderscore pigmentation &65\\
regular\textunderscore streaks &67.5\\
irregular\textunderscore streaks & 62.5\\
regular\textunderscore dots\textunderscore and\textunderscore globules&67.5\\
irregular\textunderscore dots\textunderscore and\textunderscore globules  &72.5\\
regression\textunderscore structures &70\\
dark corner &100\\
dark border &100\\
air pockets &100\\
ruler&100\\
hair&100\\
dark skin&85\\
  \end{tabular}
  \label{tab5}
\end{table}
\begin{table}[t!]

  \centering
  \scriptsize
  \renewcommand{\arraystretch}{1.5}
  \renewcommand{\tabcolsep}{6mm}
  \caption{Concept accuracy on testing set of SKINCON.}
    \vspace{-2mm}
  \begin{tabular}{c|c}
  \hline
  concept name & Acc (\%) \\
  \hline
    Papule& 65\\
    Plaque&72.5\\
    Pustule&81.82\\
    Bulla&82.57\\
    Patch&66.67\\
    Nodule&76.32\\
    Ulcer&84.38\\
    Crust&60\\
    Erosion&72.5\\
    Atrophy&57.14\\
    Exudate&86.67\\
    Telangiectasia&80\\
    Scale&73.89\\
    Scar&65.38\\
    Friable&83.33\\
    Dome-shaped&70\\
    Brown(Hyperpigmentation)&65\\
    White(Hypopigmentation)&50\\
    Purple&66.67\\
    Yellow&67.5\\
    Black&83.33\\
    Erythema&77.5\\
    dark skin&80
  \end{tabular}
  \label{tab6}
\end{table}

\subsection{More Analysis about Global Explanations}

\begin{figure}[!t]
  \begin{center}
  {\includegraphics[width=0.98\linewidth]{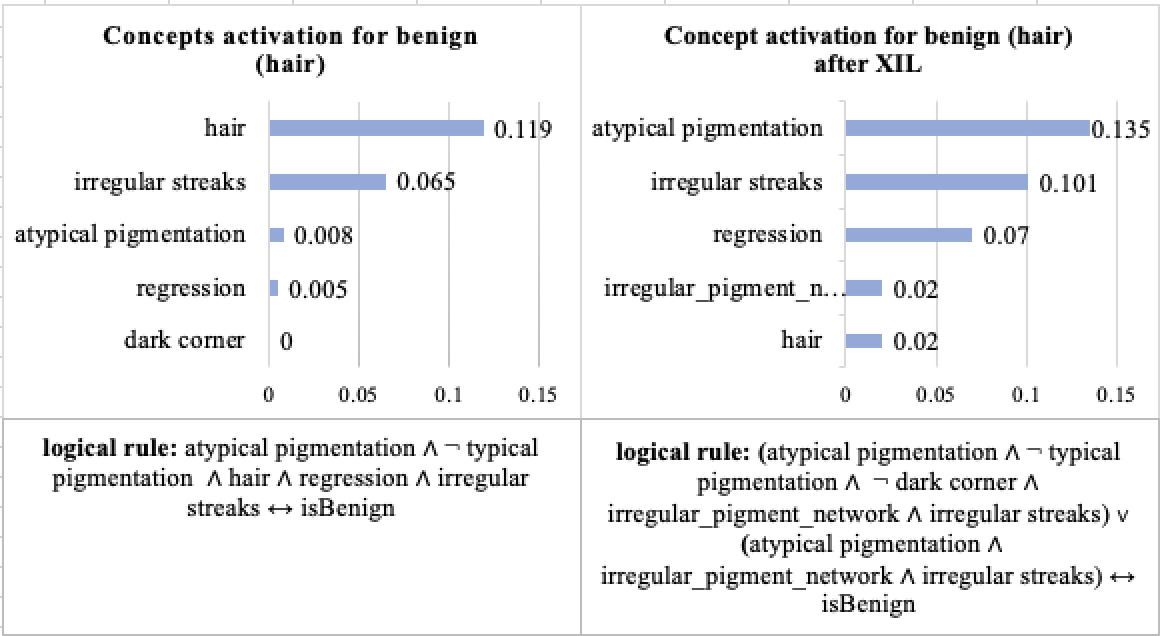}}
%   {\includegraphics[width=0.48\linewidth]{image/f5b.png}}
  \end{center}
  \vspace{-4mm}
\caption{\small The global explanation of the model's behavior on the benign (hairs) dataset of \textit{ConfDerm}. In the left figure,
either the concept activation or logical rule shows that the model is confounded by the concept of the "hair" when predicting
benign. In the right figure, after the interaction, the model relies much less on "hairs" and can predict benign based on meaningful clinical concepts.}
  \label{hair_xil}
  \vspace{-2mm}
\end{figure}

\begin{figure}[!t]
  \begin{center}
  {\includegraphics[width=0.98\linewidth]{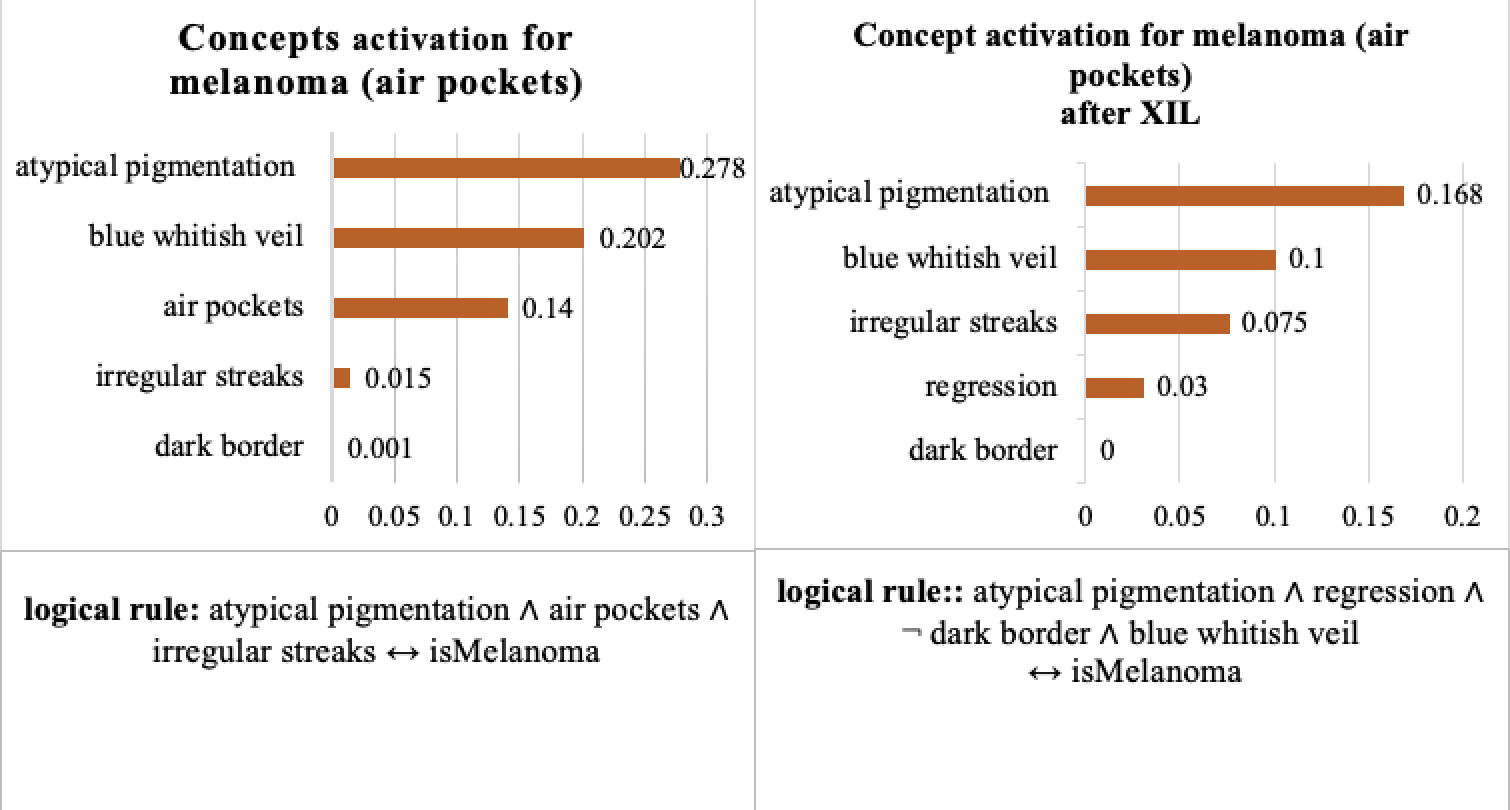}}
%   {\includegraphics[width=0.48\linewidth]{image/f5b.png}}
  \end{center}
  \vspace{-4mm}
\caption{\small The global explanation of the model's behavior on the melanoma (air pockets) dataset of \textit{ConfDerm}. In the left figure,
either the concept activation or logical rule shows that the model is confounded by the concept of the "air pockets" when predicting
melanoma. In the right figure, after the interaction, the model does not predict melanoma based on ``air pockets" and can predict benign based on meaningful clinical concepts.}
  \label{aps_xil}
  \vspace{-2mm}
\end{figure}

\noindent\textbf{Explanations of ``Rewriting Model’s Decision
in ConfDerm ":} We provide the comparison between the explanation of the model and the explanation of the model after XIL on other four datasets, including \textit{benign (dark borders), benign (rulers), benign (hairs), and melanoma (air pockets)}, as shown in Fig. \ref{db_xil}, \ref{ruler_xil}, \ref{hair_xil}, and \ref{aps_xil}. It can be seen that our XIL method can make the model focus less on confounding factors.

\begin{figure*}[!t]
   \begin{center}
      \begin{tabular}{ c@{ } c@{ } }
  {\includegraphics[width=0.98\linewidth]{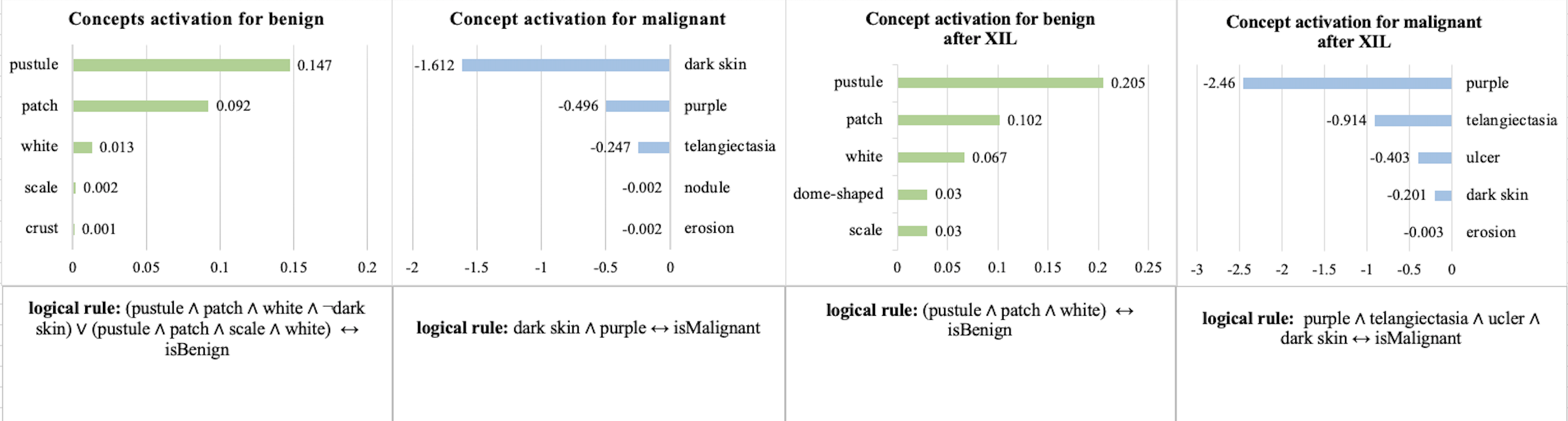}}
    \end{tabular}
   \end{center}
\caption{\small The global explanation of the model's behavior on the \textit{Fitzpatrick17k} dataset. In the two left figures,
either the concept activation or logical rule shows that the model is confounded by the concept of the dark corners when predicting
malignant. In the two right figures, after the interaction, the model relies less on "dark skin" to predict malignant, and it predicts
malignant based on meaningful clinical concepts.}
   \label{ds_xil}
\end{figure*}

\noindent\textbf{Explanations of ``Debiasing the Negative Impact
of Skin Tone":} In Fig. \ref{ds_xil}, we show the comparison between the explanation of the model and the explanation of the model after XIL on \textit{Fitzpatrick17k} dataset. It can be seen that our XIL method makes the model focus less on dark skin and can focus on meaningful clinical concepts again.

\end{document}